\documentclass{article}

\PassOptionsToPackage{numbers,compress}{natbib}

\usepackage[main, final]{neurips_2026}

\usepackage[utf8]{inputenc} 
\usepackage[T1]{fontenc}    
\usepackage{hyperref}       
\usepackage{url}            
\usepackage{booktabs}       
\usepackage{amsfonts}       
\usepackage{nicefrac}       
\usepackage{microtype}      
\usepackage{xcolor}         

\usepackage{microtype}
\usepackage{graphicx}
\usepackage{subcaption}
\usepackage{booktabs} 
\usepackage{nicefrac}
\usepackage{enumitem}
\usepackage{comment}
\usepackage{hyperref}
\usepackage{algorithm}
\usepackage{algpseudocode}
\makeatletter
\algrenewcommand\ALG@beginalgorithmic{\small}
\makeatother
\usepackage{amsmath}
\usepackage{amssymb}
\usepackage{mathtools}
\usepackage{amsthm}
\usepackage{caption}
\usepackage{wrapfig}
\theoremstyle{plain}
\newtheorem{theorem}{Theorem}[section]

\theoremstyle{definition}

\theoremstyle{remark}

\newcommand{\StateRL}{\mathcal{S}}
\newcommand{\Action}{\mathcal{A}}
\newcommand{\IS}{\mathrm{IS}}
\newcommand{\Psim}{P^\pi_{\mathrm{sim}}}
\newcommand{\Preal}{P^\pi_{\mathrm{real}}}
\DeclareMathOperator{\argmax}{\arg\max}
\DeclareMathOperator{\E}{\mathbb{E}}

\hypersetup{colorlinks,
linkcolor=blue,
citecolor=blue,
urlcolor=magenta,
linktocpage,
plainpages=false}

\title{Simulation-aided Reinforcement Learning with Control Variates}

\author{%
  Omer Hemo, Ido Sefi, Mirco Mutti, Aviv Tamar, Kfir Y. Levy  \\
  Department of Electrical and Computer Engineering \\
   Technion - Israel Institute of Technology\\
  \texttt{omer.hemo@campus.technion.ac.il} \\
}

\begin{document}

\maketitle

\begin{abstract}
Reinforcement Learning (RL) algorithms often rely on data collected from a target environment, which can be expensive and noisy. Although simulators can generate large amounts of additional data, standard simulation mixing approaches—which directly combine simulated and real experience—can degrade performance due to simulator inaccuracies, especially when they bias the solution away from the optimal policy. We propose a conservative simulation-aided learning method based on control variates, where simulated data is used solely to reduce variance in the learning process rather than directly shaping the learning target. We develop our approach for Policy-Evaluation, which is a fundamental building block of RL algorithms. Then we combine it within the actor-critic framework towards policy improvement, as well as extend it towards offline RL scenarios. Experiments on multiple MuJoCo benchmark tasks demonstrate improvements over standard simulation mixing approaches, highlighting control variates as an effective tool for simulation-aided reinforcement learning. Code is available at \url{https://github.com/OmerHemo/Simulation-Aided-Reinforcement-Learning-with-Control-Variates}.
\end{abstract}

\section{Introduction}

Reinforcement Learning~\citep[RL,][]{MMT2026RLbook} has had remarkable success on a variety of complex sequential decision problems. Nonetheless, those successes are mostly confined to settings in which almost unlimited data are available, oftentimes coming from ``perfect'' simulators of the decision process of interest, such as in videogames~\citep{mnih2013playing} or reasoning tasks~\citep{guo2025deepseek}.
In many practical scenarios instead, simulators are available, or they can be developed, but they are not ``perfect''. 
They typically cannot match the complexity of the real environment, for which they only provide a biased, approximate surrogate. This bias generates a gap between what we can learn in simulation and what we would like to achieve in the real world. Mitigating this gap is the focus of a large body of work on \emph{sim-to-real} adaptation~\citep{sadeghi2016cad2rl, tobin2017domain, matas2018sim, chebotar2019closing, akkaya2019solving, mehta2020active, du2021auto, muratore2021data, chen2021understanding, tiboni2023dropo, tiboni2023domain, wagenmaker2024overcoming, torne2024reconciling}.

Differently from previous works, here we are not interested in training robust policies in simulation for zero-shot transfer to the real world~\citep{chebotar2019closing}, nor into exploiting real world data to improve the simulator and reduce its inherent bias~\citep{tiboni2023dropo}. Instead, we \emph{embrace} the simulator bias to tackle an orthogonal research question: Can we exploit biased but abundant data coming from an ``imperfect'' simulator in combination with real-world experience to improve the real-world performance?

A representative domain to reason about the setting is as follows. We may want to train a navigation policy for a quadruped robot while interacting online with the physical world. While processing the real-world experience on our computing infrastructure, we can draw additional data from a simulator of the robot, which runs at a substantially higher frequency. While the simulator differs from the physical environment in some key parameters, such as friction, gravity,  contact forces, the effect of simulated actions may be correlated with the effect of the same action in the real world, meaning that the biased simulated data may carry some useful information to improve the real-world policy.

We formalize this setting as \emph{simulation-aided reinforcement learning}, in which the real environment and the simulator are represented by Markov decision processes (MDPs) with identical specifications except for their transition model.
A na\"ive approach~\citep{di2021sim, liu2023sim} to this setting, which we refer to as \emph{sim-mixing}, is to just feed our RL base algorithm with a mixture of the simulated and real data and to forget about the bias. This has been proved to work surprisingly well in practice, especially when the mixing parameter is carefully tuned~\citep{di2021sim}, often improving the variance relative to training on real only.
However, sim-mixing is bound to fail when the bias in simulated transitions is severe. An approach to tackle the bias is to use \emph{importance sampling} (IS) \citep{owen2013monte}, in which the probability of the simulated transitions is reweighted to match the distribution of samples from the real environment. While this results in unbiased estimates, the additional variance introduced by the weighting can grow large~\citep{huang2020importance, metelli2020importance}, possibly erasing all the benefits from exploiting the simulated data.

In this paper, we consider a \emph{control variates} approach to simulation-aided RL. In a plain statistical setting, let $X$ a quantity of interest we aim to estimate. A control variate~\citep[CV,][]{owen2013monte} is a random variable $Y$, with known expectation $\E[Y]$, that is correlated to $X$. The CV allows us to rephrase the initial estimation problem on the random variable $Z = X - (Y - \E[Y])$, which is unbiased, with provable statistical benefits~\citep{angelopoulos2023prediction}.
In our setting, $X$ stands for the quantity of interest in the real environment, $Y$ is the same quantity in simulation, for which the expectation can be obtained by drawing large number of simulated samples. With this interpretation, we develop a family of CV solutions to various simulation-aided RL settings. We corroborate the proposed CV approach both theoretically and empirically, showing that it often combines the unbiasedness of IS with the low variance of sim-mixing, while consistently outperforming or matching training on real data only, even when the bias of the simulator is severe.
Specifically, our contributions are:
\begin{itemize}[noitemsep,topsep=0pt,parsep=0pt,partopsep=0pt,leftmargin=*]
    \item We formally define the simulation-aided RL problem.
    \item We propose a control variates approach to policy evaluation in the introduced setting, for which we analyze theoretical and practical benefits.
    \item We show how improved policy evaluation can be integrated in online and offline RL algorithms, such as Soft Actor Critic~\citep{haarnoja2018soft} and Conservative Q-Learning~\citep{kumar2020conservative}. 
    \item We conduct an experimental campaign on several tasks from the MuJoCo suite, where we show that our CV approach matches or outperforms {\color{black} alternative methods for the simulation-aided RL setting, including standard sim and real mixing schemes~\citep{di2021sim}, PAR~\citep{lyu2024cross}, and VGDF~\citep{xu2023cross}.}
\end{itemize}

\section{Background}
\label{sec:background}

\textbf{Notation.}~~For a set $\mathcal{X}$, we denote $|\mathcal{X}|$ its size and $\Delta_{\mathcal{X}} := \{ p \in [0, 1]^{|\mathcal{X}|} \mid \sum_{x \in \mathcal{X}} p(x) = 1\}$ the probability simplex over it. For a random variable $X$, we denote $\E[X]$ its expectation, $\mathbb{V} [X] = \sigma^2_X$ its variance. For two random variables $X, Y$, their covariance is $\mathrm{Cov} (X, Y)$. 
{\color{black} Similarly, for random vectors we have $\mathbb{V}[X] := \E\|X - \E X\|^2_2$ and $\mathrm{Cov}(X,Y) := \E\langle X - \E X,\, Y - \E Y\rangle$, where $\|\cdot\|_p$ denotes the $\ell_p$ norm and $\langle \cdot,\cdot \rangle$ is the inner product.}

\subsection{Estimation with Biased Data}

Let $X \sim P$ a random variable distributed with $P$, and let $f$ a function of $X$ for which we aim to estimate the expectation $\E_{P} [f (X)]$. In a typical statistical setting, we have access to a set of measurements $\{f(x_i)\}_{i = 1}^n$, where $x_i$ are i.i.d. realizations of $X$. The \emph{Monte Carlo} (MC) estimator of $\E_P [f(X)]$ is then given by $\hat Y = \sum_{i=1}^n f(x_i) / n$. It is well-known that  $\mathbb{V} (\hat Y) = \sigma^2_f / n$, implying that  $\hat Y$ may encompass  a high error   when $n$ is small. In such settings, to reduce the variance of the estimate, one may turn to additional information coming from biased data.

\textbf{Importance sampling.}~~In some scenarios, we may access a large set of measurements $\{f(x_i)\}_{i = 1}^N$ from the same function $f$ on realizations $\{x_i\}_{i = 1}^N$ that are sampled i.i.d. from a  distribution $Q$ instead of the target distribution $P$. A direct MC approach on these measurements would result in a biased estimate approaching $\E_{Q} [f(X)]$ with $N \to \infty$. The \emph{importance sampling} (IS) estimator accounts for the distribution shift by computing $\hat Y_{\mathrm{IS}} = \frac{1}{N} \sum_{i = 1}^N w (x_i) f (x_i)$, where $w (x) := P(x) / Q(x)$ is the importance weight. The resulting estimate is unbiased and its variance is smaller than an MC estimator when $N \gg n$ and $P, Q$ are similar distributions~\citep{owen2013monte}.

\textbf{Control variates.}~~In other scenarios, we may access an additional set of measurements $\{h(x_i)\}_{i=1}^n$ from a function $h$ on realizations sampled i.i.d. from $X \sim P$, for which we know the expectation $\E_P(h(X))$. The function $h$ is often called a \emph{control variate}~\citep[CV,][]{owen2013monte}. We can define a control variate estimator $\hat Y_{\mathrm{CV}} = \frac{1}{n} \sum_{i=1}^n f_{\mathrm{cv}}(x_i)$ where $f_{\mathrm{cv}}(x) = f(x) - \lambda \big(h(x) - \E_P[h(X)]\big)$, such that the information from the control variate is exploited to adjust the MC estimate of $\E_P[f(X)]$ with a parameter $\lambda \in \mathbb{R}$. It is easy to see that the estimator is unbiased, while the variance of a point-estimate is given by $\sigma^2_{f_{\mathrm{cv}}} = \sigma_f^2 + \lambda^2 \sigma_h^2 - 2\lambda \mathrm{Cov}(f,h)$. The latter shows that the variance of the estimate can be reduced whenever the measurements from $f$ and $h$ are highly correlated (i.e., $\mathrm{Cov}(f,h)$ is large). Notably, the variance depends on $\lambda$, which can be tuned to minimize it as $\lambda^* = \mathrm{Cov}(f,h)/\sigma_h^2$, resulting in the point-estimate $f_{\mathrm{cv}}^*(x) := f_{\mathrm{cv}}(x)\big|_{\lambda=\lambda^*}$ with variance $\sigma^2_{f_{\mathrm{cv}}^*} = \big(1 - \frac{\mathrm{Cov}(f,h)^2}{\sigma_f^2 \sigma_h^2}\big)\sigma_f^2$, so the variance reduction grows with the absolute correlation $|\rho_{fh}| = \frac{|\mathrm{Cov}(f,h)|}{\sigma_f \sigma_h}$.

\subsection{Markov Decision Processes}

A Markov Decision Process~\citep[MDP,][]{puterman2014markov} models sequential decision problems. An MDP is defined as $M = (\StateRL, \Action, P, R, \mu, \gamma)$, where $\StateRL$ is the state space, $\Action$ is the action space, $P: \StateRL \times \Action \to \Delta_{\StateRL}$ is the transition model, $R: \StateRL \times \Action \to [0, R_\mathrm{max}]$ is the reward, $\mu \in \Delta_{\StateRL}$ is the initial state distribution, and $\gamma \in (0, 1)$ is the discount factor.
A policy $\pi : \StateRL \to \Delta_{\Action}$ describes the decision strategy of an agent interacting with the MDP according to the following protocol.
First, a state $s_0 \sim \mu$ is drawn. Then, for every step $t \geq  0$, the agent observes $s_t$, takes an action $a_t \sim \pi (s_t)$, and collects a reward $r_t = R(s_t, a_t)$, while the process transitions to the next state $s_{t + 1} \sim P(s_t, a_t)$.
{\color{black}This induces a distribution over the state-action pairs visited by $\pi$, from which the data are collected, which we denote by $P^\pi(s,a)$, and a corresponding distribution over transitions $P^\pi(s,a,s') := P^\pi(s,a)\,P(s' \mid s,a)$.}

The quality of a decision policy $\pi$ in a given state can be measured through the \emph{value function} and the \emph{state-action value function}, defined respectively as
\begin{equation*}
\begin{aligned}
V^\pi_P (s) 
&:= \E_{\pi, P} \left[ \sum_{t = 0}^{\infty} \gamma^t r_t \ \Big| \ s_0 = s\right], 
\qquad
Q^\pi_P (s, a) 
:= \E_{\pi, P} \left[ \sum_{t = 0}^{\infty} \gamma^t r_t \ \Big| \ s_0 = s, a_0 = a\right].
\end{aligned}
\end{equation*}
in which the expectation is taken over actions $a_t \sim \pi (s_t)$ and state transitions $s_{t + 1} \sim P(s_t, a_t)$, where the exponential discount ensures the sum remains finite.
Reinforcement Learning~\citep[RL,][]{MMT2026RLbook}
is a paradigm to \emph{solve} MDPs from sampled interactions, i.e., without knowledge of the MDP specification. Solving the MDP refers to two classical problems. One is called \emph{policy evaluation}, in which the goal is to estimate the value $V^\pi_P$ or $Q^\pi_P$ of a given policy $\pi$. The other is \emph{policy optimization}, where the goal is to learn a policy $\pi^* \in \argmax_{\pi} \E_{\mu} [V^\pi_P (s)]$, which is called the optimal policy.

\section{Problem Setting}
In this paper, we consider a pair of MDPs
\begin{align*}
    &M_{\mathrm{real}} = (\StateRL, \Action, P_{\mathrm{real}}, R, \mu, \gamma) &
    M_{\mathrm{sim}} = (\StateRL, \Action, P_{\mathrm{sim}}, R, \mu, \gamma)
\end{align*}
with the same specification of $\StateRL, \Action, R, \mu, \gamma$ but possibly different transition models. The MDP $M_{\mathrm{real}}$ represents the real environment, i.e., the environment of interest in which we aim to solve typical policy evaluation and optimization problems. Differently from the usual setting, here we may access an additional environment $M_{\mathrm{sim}}$ that \emph{simulates} the real environment with some approximation error.

\emph{Simulation-aided} RL is concerned with the problem of evaluating or optimizing a policy by combining the information from the real environment $M_{\mathrm{real}}$ and the simulator $M_{\mathrm{sim}}$. {\color{black}Our problem falls within the broader off-dynamics RL setting~\cite{eysenbach2020off}, which encompasses any transfer between MDPs with mismatched dynamics; our emphasis is on using simulation data to aid each policy update step in the real environment}. Clearly, the setting is trivial whenever $P_{\mathrm{sim}} \approx P_{\mathrm{real}}$ in distribution: Any RL algorithm treating simulated samples as real-world samples would be approximately optimal. The following result makes the latter statement precise.

\begin{theorem}[Lemma 10.12, \cite{MMT2026RLbook}]
\label{thr:value_discrepancy}
    Let $\pi$ a given policy and let $q \in \Delta_{\StateRL}$ the stationary state distribution induced by $\pi$ in $M_{\mathrm{real}}$. It holds
    \begin{align*}
        (1 - \gamma)^2 \E_{q} &[|V^\pi_{P_{\mathrm{real}}} (s) - V^\pi_{P_{\mathrm{sim}}} (s)|] \leq \gamma R_{\mathrm{max}} \E_{q, \pi} [\| P_{\mathrm{real}} (s,a) - P_{\mathrm{sim}} (s, a) \|_1].
    \end{align*}
\end{theorem}

In this paper, we are especially interested in problems in which $\E_{q, \pi} [\| P_{\mathrm{real}} (s,a) - P_{\mathrm{sim}} (s, a) \|_1]$ is large, which means the simulated samples shall be handled with care. In the following sections, we first focus on the problem of policy evaluation. We describe common approaches and corresponding pitfalls, as well as a method to address them. Then, we show how the proposed policy evaluation solution can bring ripple benefits to policy optimization.

\section{Simulation-aided Policy Evaluation}
Here we study the problem of policy evaluation in the simulation-aided RL setting. Let $\pi$ be the policy of which we aim to estimate the value in $M_{\mathrm{real}}$. We are provided with a dataset $D_{\mathrm{real}} =\{ (s, a, s')_{i = 1}^n \}$ taken with the policy $\pi$ in $M_{\mathrm{real}}$. With this dataset, we aim to estimate $Q^\pi_{P_{\mathrm{real}}}$ within a class of parametric models $Q_\theta$, where $\theta \in \mathbb{R}^d$. 
Typically, this is done by minimizing the squared temporal-difference (TD) error~\cite{sutton1988learning,sutton2018reinforcement}, i.e.,
\begin{equation}
\min_{\theta \in \mathbb{R}^d} \ \mathcal{L}(\theta)
:= \E_{\pi, P_{\mathrm{real}}} \big[ \delta_\theta(s,a,s')^2 \big], \quad
\text{where} \ 
\delta_\theta(s,a,s')
:= Q_\theta(s,a) - \Bigl( R(s,a) + \gamma V_\theta(s') \Bigr),
\label{TD objective}
\end{equation}
with \(V_\theta(s') := \E_{a' \sim \pi(s')} \bigl[ Q_\theta(s',a') \bigr]\). To improve stability, it is common to employ a \emph{target network} $Q^T_\theta$, which is a delayed copy of the current estimator $Q_\theta$. In particular, $Q^T_\theta(s,a) := Q^T(s,a)$, where the target parameters $\theta^T$ are updated more slowly than $\theta$, typically via an exponential moving average $\theta^T \leftarrow \tau \theta + (1 - \tau)\theta^T$ for some $\tau \in (0,1)$. The use of such slowly updated target networks is known to significantly stabilize temporal-difference learning~\citep{mnih2015human,lillicrap2015continuous,fujimoto2018addressing}. Therefore, in practice $\delta_\theta(s,a,s') = Q_\theta(s,a) - \big( R(s,a) + \gamma V^T(s') \big)$, with $V^T(s') := \E_{a' \sim \pi(s')} \bigl[ Q^T(s',a') \bigr]$.

While the problem of minimizing the squared TD error for policy evaluation is well-understood~\citep{dann2014policy}, here we are interested in studying whether a superior estimate can be obtained by exploiting an additional dataset of transitions $D_{\mathrm{sim}} =\{ (s, a, s')_{i = 1}^N \}$ taken with the policy $\pi$ in $M_{\mathrm{sim}}$.

\subsection{Common Approaches and Their Pitfalls}
\label{subsec:common methods}

Simulator-mixing approaches~\citep{di2021sim, liu2023sim} combine data from simulations and real to minimize the objective:
$
\E_{\Psim} \left[\, \delta_\theta(s,a,s')^2 \right]  +   \E_{\Preal} \left[\, \delta_\theta(s,a,s')^2 \right] 
$.
One can also control simulator bias by using a weighting coefficient $\ 0\leq\alpha\leq 1$ to get
$\alpha \E_{\Psim} \left[\, \delta_\theta(s,a,s')^2 \right]  + (1 -\alpha) \E_{\Preal} \left[\, \delta_\theta(s,a,s')^2 \right]$, as it is done in~\cite{di2021sim}. We refer to this weighted objective as \emph{regular sim-mixing} and denote the resulting Q-estimation method by \textbf{Reg-Qmix}; it serves as our main mixing baseline throughout the paper. If we recall the TD error~\eqref{TD objective}, we can see that, for every $\alpha > 0$, we  are comparing $Q_\theta(s,a)$ with biased targets. 
Importance sampling~\citep{owen2013monte} can correct the bias of the expectation as
$
   \alpha \E_{\Psim} \left[ \frac{\Preal(s,a,s')}{\Psim(s,a,s')} \delta_\theta(s,a,s')^2 \right].
$
This ratio can explode in some settings. For instance, if both the transition models are deterministic, e.g., $P_{\mathrm{real}}(s,a)=\mathbb{I}\{s'=s_1\}$ , $P_{\mathrm{sim}}(s,a)=\mathbb{I}\{s'=s_2\}$, then the IS ratio tends to zero or infinity. 

\subsection{Policy Evaluation with Control Variates}

An alternative approach to address the bias is through
control variates (CV, see Section~\ref{sec:background}). In the case of the TD error in Eq.~\ref{TD objective}, the underlying random variable is the transition tuple \( (s,a,s') \), and the corresponding function is the temporal-difference error \( \delta(s,a,s') \).

We define the real TD error and simulated Bellman error~\cite{baird1995residual,mnih2013playing} as:
\begin{align*}
    \delta^{\mathrm{real}}_\theta(s,a,s')
    &:=
    Q_\theta(s,a)
    -
    \Bigl(
        r(s,a)
        + \gamma V^T(s')
    \Bigr), 
    \\ 
    \delta^{\mathrm{sim}}_\theta(s,a)
    &:=
    Q_\theta(s,a)
    -
    \Bigl(
        r(s,a)
        + \gamma \mathbb{E}_{\tilde{s}' \sim P_{\mathrm{sim}} (\cdot | s, a)} [ V^T(\tilde{s}')]
    \Bigr).
\end{align*}
Then, we can write the CV objective
\begin{align}
    \mathcal{L}_{CV}(\theta) =  \E_{s,a,s' \sim\Preal}[\delta_{\theta}^{\mathrm{real}}(s,a,s')^2 - \lambda \delta_{\theta}^{\mathrm{sim}}(s,a)^2] + \lambda \E_{s,a \sim \Psim}[w(s,a)\delta_{\theta}^{\mathrm{sim}}(s,a)^2]
    \label{CV Loss}
\end{align}
where  $w(s,a)$ is the IS density ratio $w(s,a) = \frac{\Preal(s,a)}{\Psim(s,a)}$. Applying IS to the pair \( (s,a) \) is better than on the triplet $(s, a, s')$, since the distributions \( \Preal (s, a) \) and \( \Psim (s, a) \) are necessarily closer than the non-marginalized \( \Preal(s,a,s') \) and \( \Psim(s,a,s') \).

\textbf{Empirical Loss.} To compute \eqref{CV Loss}, we use the datasets $D_{\mathrm{real}}$ and $D_{\mathrm{sim}}$. In addition, we construct an auxiliary dataset $\tilde{D}_{\mathrm{real}} = \{(s,a,s',\{\tilde{s}'\}_{j=1}^K)\}_{i=1}^n$ by taking each $(s,a)$ pair from $D_{\mathrm{real}}$, resetting the simulator to state $s$, and executing action $a$ to generate $\tilde{s}'$. For stochastic simulators, we collect $K$ independent next-state samples for each $(s,a)$ pair.

For example, in the MuJoCo simulator, the state consists of the robot’s joint positions and velocities, which are also the inputs provided to the learned policy $\pi$. Resetting the simulator to a specified configuration of joint positions and velocities is therefore straightforward. This capability is also available in many other physics simulators, such as NVIDIA Isaac Sim \cite{isaacsim2023} and PyBullet \cite{coumans2016pybullet}.

{ \color{black}
\subsection{Analysis of the Gradient of the Empirical CV Loss}
\label{sec:gardient_analysis}

Since our goal is to optimize $\mathcal{L}_{CV}(\theta)$ using gradient-based methods, we study the bias and variance of the resulting gradient estimators.
Throughout, $\theta$ is fixed and gradients treat the target $V^T$ as a constant, as in any TD method with a target network. As is standard in covariate-shift analyses~\cite[e.g.,][]{angelopoulos2023prediction, sugiyama2007covariate, gogolashvili2023importance, liu2026unbounded}, we assume access to the density ratio $w$.
The gradient of the empirical CV loss is
\begin{equation}
\hat g_\lambda
= \frac{1}{n}\sum_{i=1}^n \Big[\nabla_\theta \delta^{\mathrm{real}}_\theta(x_i)^2 - \lambda\, \nabla_\theta \delta^{\mathrm{sim}}_\theta(x_i)^2\Big]
+ \frac{\lambda}{N}\sum_{j=1}^N w(\tilde s_j,\tilde a_j)\,\nabla_\theta \delta^{\mathrm{sim}}_\theta(\tilde x_j)^2,
\label{eq:cv_grad}
\end{equation}
where $x_i \in \tilde D_{\mathrm{real}}$, $\tilde x_j \in D_{\mathrm{sim}}$, and $\delta^{\mathrm{sim}}_\theta$ uses the average of $V^T$ over the available simulated next states ($K$ for $\tilde D_{\mathrm{real}}$, one for $D_{\mathrm{sim}}$). Its behaviour is governed by the \emph{expected} real and simulated gradients at a state--action pair,
\begin{align*}
b(s,a) &:= 2\big(Q_\theta(s,a) - r(s,a) - \gamma\, \E_{s' \sim P_{\mathrm{real}}(s,a)} V^T(s')\big)\nabla_\theta Q_\theta(s,a), \\
\tilde b(s,a) &:= 2\big(Q_\theta(s,a) - r(s,a) - \gamma\, \E_{\tilde s' \sim P_{\mathrm{sim}}(s,a)} V^T(\tilde s')\big)\nabla_\theta Q_\theta(s,a),
\end{align*}
together with the \emph{transition noise} and the variance of the importance-weighted term,
\begin{align*}
\sigma_T^2 &:= 4\gamma^2 \E_{s,a\sim\Preal}\Big[\|\nabla_\theta Q_\theta(s,a)\|^2_2\, \mathbb{V}_{s'\sim P_{\mathrm{real}}(s,a)}\big[V^T(s')\big]\Big], \\
\sigma_{\mathrm{IS}}^2 &:= \mathbb{V}_{\tilde x \sim \Psim}\big[w(\tilde s,\tilde a)\, \nabla_\theta\delta^{\mathrm{sim}}_\theta(\tilde x)^2 \big],
\end{align*}
the simulator counterpart $\tilde\sigma_T^2$ of $\sigma_T^2$ (same expression with $P_{\mathrm{sim}}$) and the simulator value mismatch
\begin{equation*}
    \Delta^2_{V^T} := \E_{s,a\sim\Preal}\Big[\big(\E_{s'\sim P_{\mathrm{real}}(s,a)}V^T(s')
     - \E_{\tilde s'\sim P_{\mathrm{sim}}( s,a)}V^T(\tilde s')\big)^2\Big]
\end{equation*}

We can characterize the bias and variance of the CV loss gradient estimator as follows.
\begin{theorem}[Gradient bias and variance]
\label{thm:gradient_analysis}
Assume access to the density ratio $w$ and let the samples in $\tilde D_{\mathrm{real}}$ and $D_{\mathrm{sim}}$ be drawn independently. For every $\lambda \in \mathbb{R}$ and $K \geq 1$:
\begin{enumerate}[label=(\roman*), leftmargin=*, nosep]
    \item $\E[\hat g_\lambda] = \E_{s,a\sim\Preal}[b(s,a)] = \E [\hat g_0]$, which is the gradient of the TD objective~\eqref{TD objective};
    \item $\displaystyle \mathbb{V}[\hat g_\lambda] = \frac{1}{n}\Big(\mathbb{V}\big[b - \lambda \tilde b\big] + \sigma_T^2 + \frac{\lambda^2\tilde\sigma_T^2}{K}\Big) + \frac{\lambda^2}{N}\sigma_{\mathrm{IS}}^2$, \quad with $b, \tilde b$ evaluated at $(s,a)\sim\Preal$;
    \item $\mathbb{V}[\hat g_{\lambda^\star}] \;\le\; \,\mathbb{V}[\hat g_0]$ \quad for the variance minimizer $\lambda^\star = \frac{\mathrm{Cov}(b,\tilde b)}{\mathbb{V}[\tilde b] + \tilde\sigma_T^2/K + n\,\sigma_{\mathrm{IS}}^2/N}$;
    \item $\mathbb{V}\big[b-\lambda\tilde b\big] \le 2(1-\lambda)^2\,\mathbb{V}[b] + 8\lambda^2\gamma^2G^2\,\Delta^2_{V^T},$ \quad with $\|\nabla_\theta Q_\theta(s,a)\|_2 \le G$ for all $(s,a)$.
\end{enumerate}
\end{theorem}

\textit{Proof.} See Appendix~\ref{app:gradient_analysis_proof}.

Theorem~\ref{thm:gradient_analysis} has four parts. Part {\em (i)} shows that the estimator of the CV gradient is unbiased. Part {\em (ii)} reports an exact expression of the variance. Setting $\lambda = 0$ recovers real-only TD, whose variance $(\mathbb{V}[b] + \sigma_T^2)/n$ has two sources: the randomness of the real next state ($\sigma_T^2$) and which state-action pairs happen to be in $D_{\mathrm{real}}$ ($\mathbb{V}[b]$). Because simulated next states are drawn independently of real ones, the simulator cannot reduce $\sigma_T^2$, it acts on $\mathbb{V}[b]$ only. There, $\mathbb{V}[b - \lambda\tilde b] = \mathbb{V}[b] - 2\lambda\,\mathrm{Cov}(b,\tilde b) + \lambda^2\mathbb{V}[\tilde b]$ is exactly the classical control-variate variance of Sec.~\ref{sec:background} with $f = b$ and $h = \tilde b$: The simulator's expected Bellman gradient is a control variate for the real one. The last two terms are the price of estimating $\E[\tilde b]$ and the simulator expectation from finitely many samples, and vanish in the abundant-simulation regime $N, K \to \infty$. 
Part {\em (iii)} shows that, with a well-chosen $\lambda$, the CV gradient reduces the variance of real-only TD when $\mathrm{Cov}(b,\tilde b) \neq 0$. Part {\em (iv)} ties this gain to simulator quality: $\lambda$ removes state-action variance at rate $(1-\lambda)^2$ and pays for the simulator value mismatch at rate $\lambda^2$. This is the key difference from sim-mixing. By Theorem~\ref{thr:value_discrepancy}, a mismatched simulator biases the sim-mixing solution. Here, the mismatch enters only the variance, scaled by $1/n$ and controlled by $\lambda$.

}

\subsection{Practical Implementation of the CV Method}
\label{sec:implementation-details}
\begin{wrapfigure}{r}{0.48\textwidth}
\vspace{-10pt} 

\begin{minipage}{0.46\textwidth}
\begin{algorithm}[H]
\caption{CV-Qmix}
\label{alg:q_estimation}
\begin{algorithmic}[1]
\item \textbf{Input:}
\item \quad $D_{\mathrm{real}}, D_{\mathrm{sim}}$
\item \quad Classifier $P_\psi(x\in \text{real} \mid x)$
\item \quad Learning rate $\xi$
\item \quad Batch size $B$
\item \quad EMA decay rates $0\leq \tau_{1,2} \leq 1$
\item \textbf{Initialize:}
\item \quad $\theta \in \mathbb{R}^d$, $Q^T = Q_\theta$, $\lambda = 0$
\item \quad Compute $\tilde{D}_\mathrm{real}$
\item \textbf{for} {$i = 1$ to $T$}
    \item \quad $\{ x \sim \tilde{D}_{\mathrm{real}} \}_{i = 1}^{B},\{ x \sim D_{\mathrm{sim}} \}_{i = 1}^{B} $
    \item \quad Compute $L_{CV}(\theta)$ from (\ref{CV empiric loss})
    \item \quad Update $\theta \gets \theta - \xi \nabla L_{CV} (\theta)$
    \item \quad Update $\theta^T \leftarrow \tau_1 \theta^T + (1 - \tau_1) \theta$
    \item \quad Update $\lambda \leftarrow \tau_2 \lambda + (1-\tau_2) \hat\lambda$
\item \textbf{end for}
\item \textbf{Return} $Q_\theta$
\end{algorithmic}
\end{algorithm}
\end{minipage}
\vspace{-10pt}
\end{wrapfigure}
After the theoretical analysis of the CV loss we now look into deriving a fully practical algorithm to optimize the objective \eqref{CV Loss}. We summarize its main steps in Algorithm~\ref{alg:q_estimation}, which we refer to as \textbf{CV-Qmix} in the remainder of the paper.

In cases where we do not have access to $w(s,a)$, it can be estimated using a subset of $D_\mathrm{real}, D_\mathrm{sim}$. A large body of works have shown that this can be done in practice by training a classifier (e.g., \cite{menon2016linking}) where $w_\psi(s,a) :=\frac{P_\psi(x \in \mathrm{real} \mid x)}{P_\psi(x \in \mathrm{sim} \mid x)}\approx\frac{P^\pi_{\mathrm{real}}(s,a)}{P^\pi_{\mathrm{sim}}(s,a)}$.

The classifier is trained by minimizing the cross-entropy 
$
\label{eq:ce_loss}
\mathcal{L}_{\mathrm{CE}}(\psi)
=
- \mathbb{E}_{x \sim D_{\mathrm{real}}}
\big[ \log P_\psi(y=1 \mid x) \big]
-
\mathbb{E}_{x \sim D_{\mathrm{sim}}}
\big[ \log P_\psi(y=0 \mid x) \big]
$.

In addition, we observe that the CV loss in Eq.~\ref{CV Loss} can become unstable when state--action pairs that appear predominantly in the real dataset are overfit by the value function \( Q_\theta(s,a) \), leading to excessively negative contributions in the first expectation. To mitigate this issue, we apply zero masking, $\mathbb{I}_m := \mathbb{I}\{ w_\psi(x) < \beta_w\}$ to samples with large importance weights, effectively excluding simulator samples for which \( w_\psi(s,a) > \beta_w \). The masked loss:
\begin{align}
\label{CV empiric loss}
    L_{CV}(\theta) = \frac{1}{n}\sum_{i=1}^n\Big[ \delta_{\theta}^{\mathrm{real}}(x_i)^2 - \mathbb{I}_m\,\lambda\, \delta_{\theta}^{\mathrm{sim}}(x_i)^2\Big]
    + \frac{1}{N} \sum_{j=1}^N \Big[ \mathbb{I}_m\,\lambda\, w_\psi(\tilde s_j,\tilde a_j)\,\delta_{\theta}^{\mathrm{sim}}(\tilde x_j)^2 \Big]
\end{align}

{
\color{black}
Importantly, the estimator is still unbiased (Apx.~\ref{app:mask_bias_proof}) and, since the mask is a function of $(s,a)$ only, Theorem~\ref{thm:gradient_analysis} holds verbatim for the masked estimator, with $\tilde b$ and the simulator terms multiplied by $\mathbb{I}_m$. The mask also controls the importance-sampling term: Since $\E_{\Psim}\big[\mathbb{I}_mw^2\|\nabla_\theta(\delta^{\mathrm{sim}}_\theta)^2\|^2_2\big] = \E_{\Preal}\big[\mathbb{I}_mw\,\|\nabla_\theta(\delta^{\mathrm{sim}}_\theta)^2\|^2_2\big]$, for the mask computed with the true $w$ we get $\sigma_{\mathrm{IS}}^2 \leq \beta_w\, \E_{\Preal} \big[\|\nabla_\theta (\delta^{\mathrm{sim}}_\theta)^2\|^2_2\big]$, whereas without masking it can be arbitrarily large when the two state-action distributions overlap poorly.

We tune the mixing coefficient $\lambda$ using the control variate formulation at the loss level rather than at the gradient level, as a cheap scalar surrogate of the $\lambda^\star$ in Theorem~\ref{thm:gradient_analysis}{\em(iii)}. We find this approach to be effective for gating simulated data. When updating the target network $Q^T(s,a)$, we also update $\lambda$ according to:
$\hat\lambda = \mathrm{Cov}\big((\delta^{\mathrm{real}}_\theta)^2, (\delta^{\mathrm{sim}}_\theta)^2\big) / \mathbb{V}\big((\delta^{\mathrm{sim}}_\theta)^2\big)$.
}

\section{Simulation-aided Policy Optimization}

After analyzing the policy evaluation problem, we now shift the focus to policy optimization in simulation-aided RL. In this setting, we consider a class of parametric policies $\pi_{\phi}$, where $\phi \in \mathbb{R}^z$. The typical goal is to learn a policy
\begin{equation}
\label{eq:policy_optimization}
    \phi^* \in \argmax_{\phi \in \mathbb{R}^z} \E_{\mu} [V^{\pi_\phi}_{P_{\mathrm{real}}} (s)]
\end{equation}
from sampled interactions with the MDP $M_{\mathrm{real}}$. The interactions may come \emph{online} by querying the MDP with a policy of our choice, or \emph{offline}, in which data are gathered beforehand by running one or multiple policies in the MDP.

In modern algorithms, the optimization of the objective \eqref{eq:policy_optimization} is typically done by iterative gradient ascent updates over the current policy parameters $\phi_t$ or the current value parameters $\theta_t$, from which the policy parameters are recovered later. In both settings, the policy evaluation step is crucial to either estimate the policy gradient~\citep{schulman2017proximal, haarnoja2018soft} or as the main target of the optimization~\citep{mnih2013playing, kumar2020conservative}.

In the simulation-aided setting, we further access transitions coming from an imperfect simulator $M_{\mathrm{sim}}$, which we may use to improve the gradient estimates at each step.
Especially, we aim to test whether our simulation-aided policy evaluation method can bring benefits to the corresponding policy optimization problem as well. We detail below how to incorporate Algorithm~\ref{alg:q_estimation} as a policy evaluation building block in common actor-critic algorithms for the online and offline settings, respectively.

\subsection{Online Policy Optimization}

For the online setting, we use Soft Actor-Critic~\citep[SAC,][]{haarnoja2018soft} as a representative approach, although our Q estimation method can be easily incorporated in other popular algorithms, such as PPO~\citep{schulman2017proximal} or TD3~\citep{fujimoto2018addressing}. To extend SAC to the simulation-aided setting, we just need to incorporate our Q estimation procedure in the critic update step and an additional sampling routine to collect fresh data from $M_{\mathrm{sim}}$ with the current policy at each iteration.
Crucially, SAC makes use of entropy regularization, so that we need to substitute in Algorithm~\ref{alg:q_estimation} the standard TD error~\eqref{TD objective} with a \emph{soft Bellman error}
$
    \delta^{\mathrm{soft}}_\theta (s, a, s') := Q_\theta (s,a) - y
$
where the target $y$ is entropy-regularized with some $\alpha > 0$,
$$
    y = R(s, a) + \gamma \E_{a' \sim \pi_{\phi} (s')} \Big[ Q^T (s', a') - \alpha \log \pi_{\phi} (a' | s') \Big].
$$

\subsection{Offline Policy Optimization}

Collecting data online with the current policy is often impractical in safety-critical and expensive domains, such as robotic manipulation or autonomous driving. In those settings, one would rather collect the data once and then iterate on the fixed dataset to learn a good policy. In the simulation-aided setting, we assume to have a pre-collected dataset from both the environment $M_{\mathrm{real}}$ and its simulator $M_{\mathrm{sim}}$.
Similarly as in the previous section, we want to extend popular offline RL algorithms with our Q estimation procedure. Here we chose Conservative Q-Learning~\citep[CQL,][]{kumar2020conservative} as a representative example.
Extending our method to the offline setting introduces two additional challenges. First, if the real and simulated datasets, $D_{\mathrm{real}}$ and $D_{\mathrm{sim}}$, are collected under different policies, then the induced state-action distributions $P^{\pi_1}_{\mathrm{sim}}$ and $P^{\pi_2}_{\mathrm{real}}$ may have limited overlap. As a result, the masking mechanism and density ratio term $w(s,a)$ (Eq.~\ref{CV empiric loss}) may discard a large portion of the data.

The high-overlap condition is naturally satisfied when both datasets are generated by the same behavior policy. While this coupling is straightforward to enforce in the online setting, it is more challenging offline, where real-world data may be collected under an unknown policy. In our current formulation, we assume access to the behavior policy. A practical instance of this setting arises when the behavior policy is trained in simulation and then used to collect real-world data.

When incorporating simulated data into CQL, a second challenge arises from dataset imbalance. In particular, when the amount of simulated data is much larger than the real dataset (\( N \gg n \)), the CQL regularization term can be dominated by simulated
transitions, effectively under-regularizing the real data. To address this issue, we apply CQL regularization separately to the simulated and real datasets.

\section{Experimental Results}
\label{sec:experiments}

We evaluate our method on standard MuJoCo continuous-control benchmarks \cite{todorov2012mujoco}, and on a custom MuJoCo robotics task, \textit{Puck Sliding}, inspired by \cite{plappert2018multi} (see Appendix \ref{sec:Puck-sliding} for details). To simulate the sim-to-real gap, we construct distinct real and simulated environments by perturbing physical parameters such as gravity, contact friction, joint damping, body part lengths, and actuator strength.

\textbf{Baselines.}~~In all of the domains, we evaluate our approach, \textbf{CV-Qmix}, against several baselines. The most relevant are \textbf{Reg-Qmix}, a standard approach to mix sim and real data without correction (see Sec.~\ref{subsec:common methods}), closely inspired by~\citep{di2021sim}, and \textbf{Real}, which is a standard TD critic operating on real data only, corresponding to CV-Qmix with $\lambda = 0$. While these are policy evaluation methods, in policy optimization experiments we use the same algorithms' names to refer to actor-critic methods in which the critic is implemented by CV-Qmix, Reg-Qmix, Real, respectively. The actor-critic backbone is SAC~\citep{haarnoja2018soft} for online policy optimization and CQL~\citep{kumar2020conservative} for offline policy optimization. In online policy optimization, we further compare against \textbf{PAR}~\cite{lyu2024cross} and \textbf{VGDF}~\cite{xu2023cross}, two strong baselines from the ODRL benchmark~\cite{lyu2024odrl} (see Section~\ref{sec:related} for details on how they work).

\begin{figure}[t]
    \centering
    \includegraphics[width=1\linewidth]{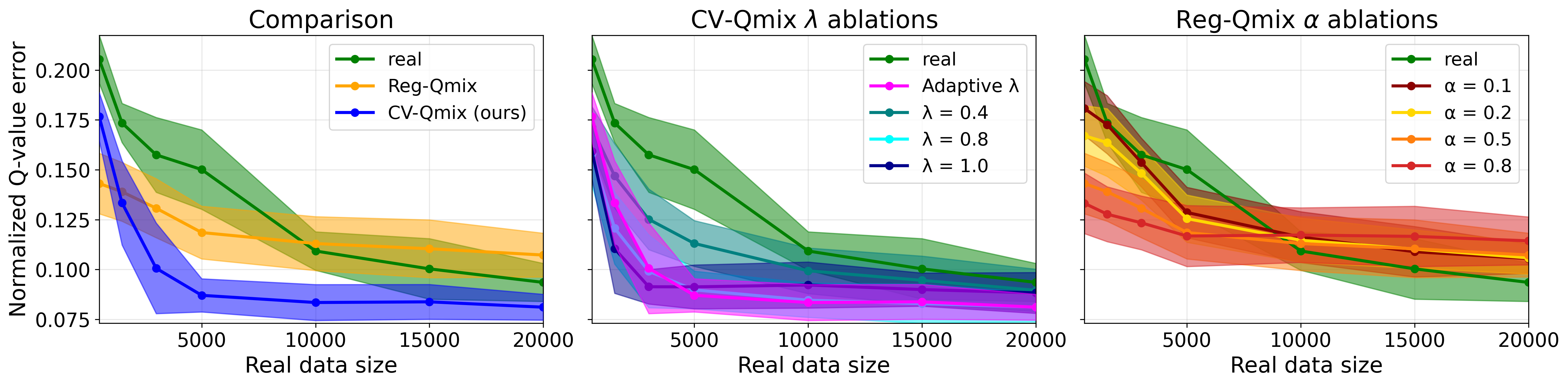}
    \caption{\textbf{Simulation-aided policy evaluation on the Humanoid task.} Simulation parameters (gravity tilt and magnitude) are perturbed relative to the real environment. \textbf{Left:} Normalized Q-value error of each mixing method as a function of the real-data budget. \textbf{Middle:} Effect of $\lambda$ on CV-Qmix, including the adaptive choice of $\lambda$. \textbf{Right:} Effect of $\alpha$ on Reg-Qmix. In all plots, the real-only estimator is shown for reference. Curves show the mean and shaded regions $\pm$1 std over 10 seeds.}
    \label{fig:pol-eval-exp}
\end{figure}

\textbf{Experiments.}~~We first evaluate CV-Qmix on policy evaluation, the task it is designed to address, against Reg-Qmix and Real (Section~\ref{sec:exp-pe}, Figure~\ref{fig:pol-eval-exp}). We further ablate the CV coefficient $\lambda$ of CV-Qmix and the mixing coefficient $\alpha$ of Reg-Qmix (Figure~\ref{fig:pol-eval-exp} center, right). We then test whether these gains carry over to online and offline policy optimization (Section~\ref{sec:exp-po}, Figure~\ref{fig:off-dynamics}~\ref{fig sample efficiency}). Throughout, we vary the \emph{sim gap}, i.e., how far the simulator's dynamics parameters are shifted from those of $M_{\mathrm{real}}$: a larger gap corresponds to a larger shift and therefore a more biased simulator (exact values in Apx~\ref{app: sim parameters}). Additional results on robustness to increasing sim gaps and on the values of the adaptive $\lambda$ are reported in Apx.~\ref{app:robustness} and~\ref{app:lambda}.

\subsection{Policy Evaluation Experiments}
\label{sec:exp-pe}
We evaluate on the MuJoCo Humanoid task, where $M_{\mathrm{real}}$ uses the default dynamics and $M_{\mathrm{sim}}$ perturbs the gravity magnitude and tilt (Appendix~\ref{app: sim parameters}). We report the \emph{normalized Q-value error}: the mean absolute error between the Monte Carlo Q-values predicted by the methods and ground-truth Q-values, after min-max normalizing both by the range of the ground-truth returns. To obtain ground-truth values, we construct a test set of state--action pairs $(s,a)$ and estimate the Q-value of each pair under the evaluation policy by averaging the returns of 10 rollouts from $(s,a)$ in $M_{\mathrm{real}}$.

Figure~\ref{fig:pol-eval-exp} (left) shows that CV-Qmix outperforms the real-only estimator at every real-data budget. At the smallest budget, Reg-Qmix attains the lowest error, but CV-Qmix overtakes it beyond roughly 2{,}000 samples and, unlike Reg-Qmix, keeps improving and plateaus at a lower error as real data grows. The Reg-Qmix ablation (right) shows why naive mixing falls short. A high simulated-data ratio ($\alpha=0.8$) reduces error when real data is scarce but plateaus above the real-only baseline, whereas a low ratio ($\alpha=0.2$) tracks the baseline closely: more simulated data trades variance for bias. In contrast, CV-Qmix uses simulated data to reduce the variance of the real-data estimator while correcting for the simulator's bias. As Theorem~\ref{thm:gradient_analysis} shows, the variance is quadratic in $\lambda$ and therefore minimized at an intermediate value. In this experiment, $\lambda=0.8$ performs best (middle), and the adaptive estimate of $\lambda$ (Section~\ref{sec:implementation-details}) comes close to this best fixed value.

\subsection{Policy Optimization Experiments}
\label{sec:exp-po}

We evaluate how improved policy evaluation can benefit policy optimization, both online and offline, in  variety of MuJoCo domains, Ant, Walker, Hopper, and Puck Sliding.

\begin{figure}
    \centering
    \includegraphics[width=1\linewidth]{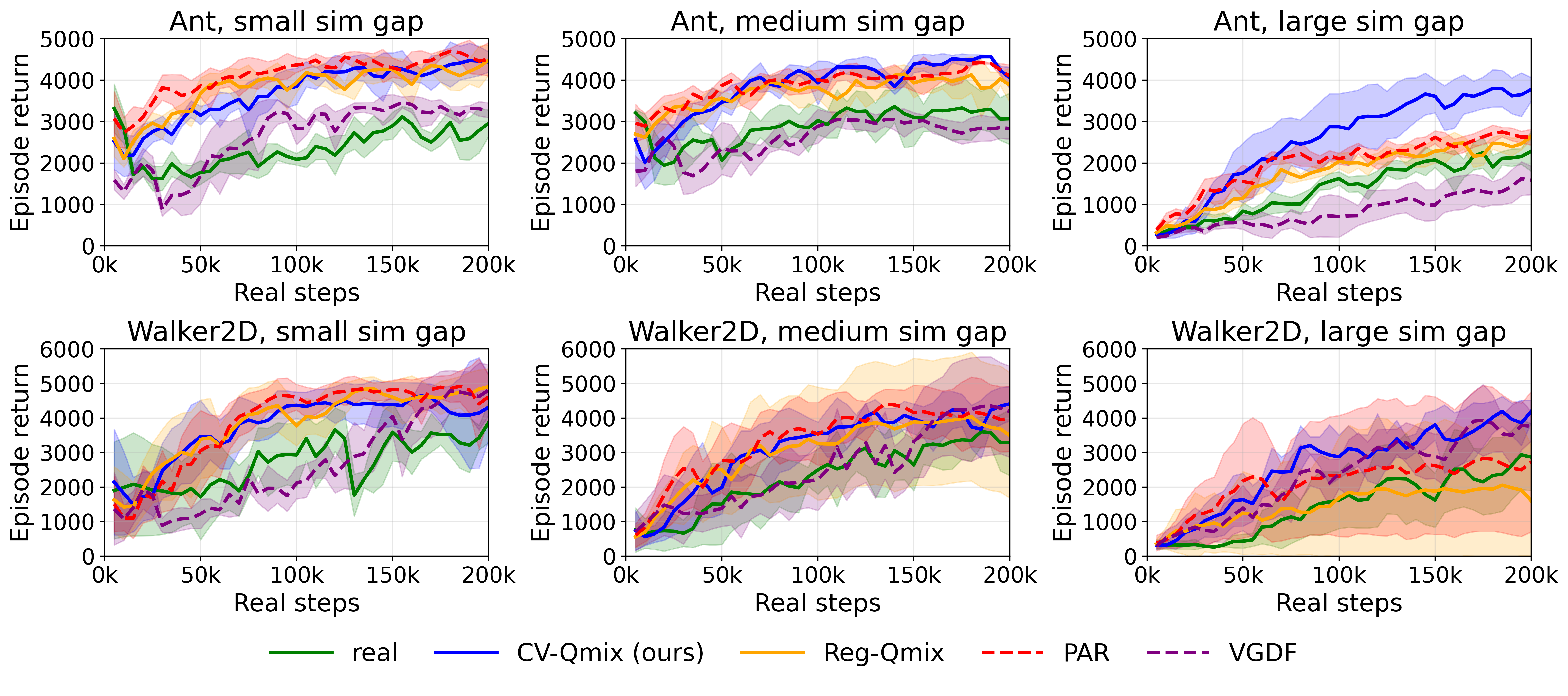}
    \caption{\textbf{Policy finetuning with biased simulation data mixing.} Comparing CV-Qmix (ours) and Reg-Qmix to other off-dynamics methods on Ant and Walker2D tasks across different simulation to real gaps. The policies were pre-trained on the simulation and then transferred to the real environment for fine-tuning with mixed simulation and real data. Each plot shows the mean $\pm$ std of 5 seeds.}
    \label{fig:off-dynamics}
\end{figure}

\textbf{Online policy optimization.}~~Figure~\ref{fig:off-dynamics} shows the results. In this experiment, all policies are pre-trained in simulation with SAC, and then fine-tuned using both simulated and real data (except the Real baseline) with SAC equipped with the corresponding critic. The simulator and real environment differ in their dynamics parameters at three gap levels (small, medium, and large; see Appendix~\ref{app: sim parameters} for dynamics shift details). The plots report the performance of the fine-tuning stage. As we can see, CV-Qmix performs better or on par with the most competitive baseline across all the gap levels, with the benefit of CV-Qmix approach increasing with larger sim gaps. On Ant, CV-Qmix is comparable to other baselines at small and medium gap levels, but on the large gap it clearly outperforms all the baselines, with the margin growing with the gap. On Walker2D, the methods are comparable at small and medium gaps, while at a large gap Reg-Qmix and PAR degrade and CV-Qmix remains the best together with VGDF. These results are consistent with bias correction becoming important as the gap grows. It is interesting to take a closer look at ODRL baselines. PAR closely tracks Reg-Qmix: both train on mixed data, and PAR's reward penalty does not remove the bias of simulated TD targets. VGDF avoids this bias, by filtering out simulation samples, but filtering too much simulated samples costs sample efficiency.

\begin{figure*}[t]
    \centering
    \includegraphics[width=1\textwidth]{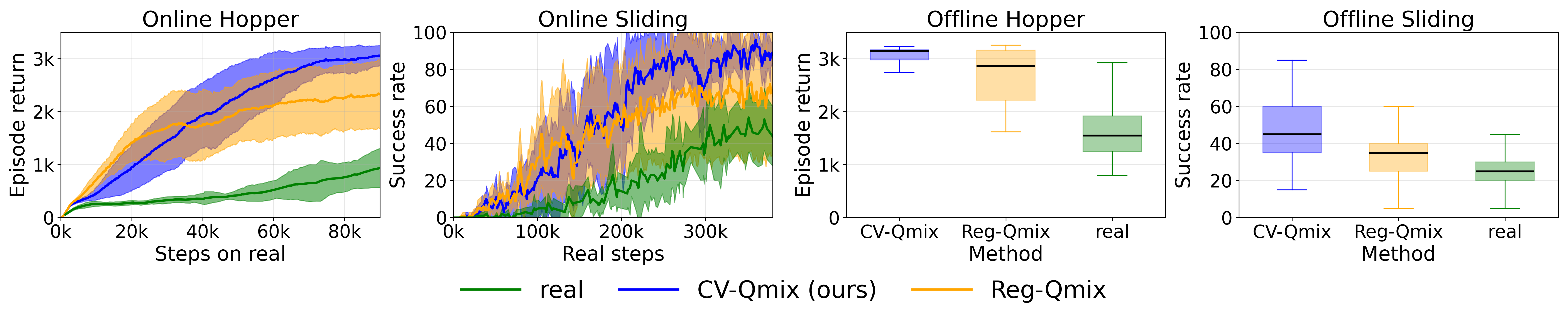}
    \caption{\textbf{Policy optimization with mixed simulated data on Hopper and Puck Sliding}. Simulator parameters (gravity, friction, and body part lengths) are perturbed relative to the real environment. The plots on the left show online optimization (mean $\pm$ std over 10 seeds), while the plots on the right show offline optimization (box plots over 10 seeds).}
    \label{fig sample efficiency}
\end{figure*}

In figure~\ref{fig sample efficiency} (left), we test whether simulated data enables more policy updates, performing $10\times$ more updates per real interaction. $M_{\mathrm{real}}$ uses the default MuJoCo dynamics and $M_{\mathrm{sim}}$ uses perturbed dynamics (Appendix~\ref{app: sim parameters}). Figure~\ref{fig sample efficiency} compares Reg-Qmix and CV-Qmix (ours) against standard SAC (Real), which uses only real samples. Both mixing methods outperform the baseline, but Reg-Qmix plateaus earlier and has higher variance. We attribute the higher variance to the instability of jointly minimizing the real and simulated expectations (Sec.~\ref{subsec:common methods}), which is sensitive to the real data collected during training; CV-Qmix reduces the variance of the real expectation, giving more stable optimization. We attribute the earlier plateau and lower asymptotic performance of Reg-Qmix to simulator bias, which can lead the policy toward actions that are suboptimal in the real environment.

\textbf{Offline policy optimization.}~~In the offline setting (Figure~\ref{fig sample efficiency}, right), we first train a behavioral policy in the simulator with SAC until it plateaus, then use it to collect $D_{\mathrm{real}}$ and $D_{\mathrm{sim}}$, with substantially more simulated than real data (Appendix~\ref{app:Training parameters}). We run CQL with a CV-Qmix, Reg-Qmix, and real only critic with those data. Since offline policies cannot be evaluated by interacting with the real environment during training, early stopping and model selection are difficult~\cite{levine2020offline, paine2020hyperparameter}; we therefore report performance averaged over the last 50k of 200k update steps, summarized as box plots. CV-Qmix outperforms both Reg-Qmix and real-only training on the Hopper and Sliding tasks. These results highlight a further advantage of CV-Qmix in the offline setting: by reducing the variance of the learning process (most visibly on Hopper), it makes performance less sensitive to the choice of stopping point, which matters when early stopping and model selection are difficult.

\section{Related Work}
\label{sec:related}

\textbf{Off-dynamics RL.}~~ Our setting falls within off-dynamics RL~\cite{eysenbach2020off}, where a source and a target MDP differ only in their transition dynamics, and access to the target is limited while the source is cheap to query. Existing methods handle the dynamics mismatch per transition, by correcting, penalizing, or discarding source samples: DARC~\cite{eysenbach2020off} corrects source rewards using domain classifiers, PAR~\cite{lyu2024cross} penalizes them by the representation mismatch across domains, and VGDF~\cite{xu2023cross} discards source transitions whose value targets disagree with those expected under the target dynamics, a criterion closely related to our $\Delta^2_{V^T}$ term. In contrast, we use simulation data to aid each policy update in the real environment: rather than discarding simulated samples or modifying their rewards, we exploit their correlation with real ones and correct their bias at the policy-evaluation level, through a control-variate formulation of the TD loss~\eqref{TD objective}. We compare against PAR and VGDF in Section~\ref{sec:experiments}.

\textbf{Control variates in RL.}~~Control variates (CV) are a standard variance-reduction tool in RL, e.g., baselines and action-dependent critics in policy-gradient methods~\cite{gu2016q}. In off-policy evaluation, the doubly robust (DR) estimator~\cite{jiang2016doubly} combines importance-weighted returns with a learned value model as a CV. DR is designed to debias off-\emph{policy} data, while we debias off-\emph{dynamics} data. Moreover, DR applies the CV directly to the policy-evaluation objective, while we apply it to the squared TD error. Statistically, our approach is closer to prediction-powered inference~\cite{angelopoulos2023prediction}, which combines many biased model predictions with a few ground-truth samples to obtain unbiased, lower-variance estimates; here, simulated transitions play the role of the biased predictions.

\section{Conclusion}
In this paper, we introduced the simulation-aided reinforcement learning and a control variates approach that exploits biased simulator data to reduce variance without negatively biasing the performance on the real environment. Applying control variates to the squared TD error with a stable importance-sampling correction enables seamless integration into both online and offline actor–critic methods. Theory shows that variance reduction improves as the simulator more closely matches the real dynamics. Experiments on MuJoCo tasks demonstrate improved sample efficiency and robustness over standard simulation mixing, especially large under dynamics mismatch.

Notably, our approach is entirely focused on policy evaluation, which is then incorporated into policy optimization algorithms as a building block. Future works may extend our control variates approach to the policy gradient estimate, bringing variance reduction in both the actor-critic components.
Another future direction may consider a similar control variates solution to the model-based RL approach, in which we aim to estimate the transition dynamics of the real environment by exploiting information coming from multiple biased simulators.

Finally, our approach requires resetting the simulator to visited real states, and therefore does not directly apply in partially observable settings where the real states cannot be fully accessed~\cite{kaelbling1998planning, thrun2005probabilistic}. One natural resolution is to adapt our method to such settings by constructing an estimator that predicts the state from partial observations, in the spirit of ~\cite{poupart2000value, roy2004belief, corso2023betazero}, and to then reset the simulator according to this state-estimation. Curiously, \cite{kumar2021rma} show that training such a predictor over the simulator enables to perform well on real partially-observable environments. While this is only one possible extension, it will be interesting to explore other alternative directions.

\section*{Acknowledgments}
OH and KYL acknowledge support from the Israel Science Foundation (grant No. 3109/24), and were also partially supported by the Israel PBC-VATAT. IS, MM and AT were funded by the European Union (ERC, Bayes-RL, 101041250).
Views and opinions expressed are however those of the author(s) only and do not necessarily reflect those of the European Union or the European Research Council Executive Agency (ERCEA). Neither the European Union nor the granting authority can be held responsible for them.

\bibliography{biblio}
\bibliographystyle{plainnat}

\newpage
\appendix
\onecolumn
\section{Proofs}

\subsection{Proof of Theorem~\ref{thr:value_discrepancy}}

To prove the result, we write
\begin{align}
    &\E_q [| V_{P_{\mathrm{real}}}^\pi (s) - V_{P_{\mathrm{sim}}}^\pi (s) |] \nonumber \\
    &\;\leq \gamma \E_{q, \pi} \left[ \left| \sum_{s' \in \StateRL} P_{\mathrm{real}} (s' | s, a) V_{P_{\mathrm{real}}}^\pi (s') -  \sum_{s' \in \StateRL} P_{\mathrm{sim}} (s' | s, a) V_{P_{\mathrm{sim}}}^\pi (s')  \right| \right] \label{eq_1} \\
    &\;= \gamma \E_{q,\pi} \left[ \left| \sum_{s' \in \StateRL} P_{\mathrm{real}} (s' | s, a) \Big( V_{P_{\mathrm{real}}}^\pi (s') -  V_{P_{\mathrm{sim}}}^\pi (s') \Big) +  \sum_{s' \in \StateRL} V_{P_{\mathrm{sim}}}^\pi (s') \Big( P_{\mathrm{real}} (s' |s,a) -P_{\mathrm{sim}} (s' | s, a) \Big) \right| \right] \label{eq_2} \\
    &\; \leq \gamma \E_q \left[ | V_{P_{\mathrm{real}}}^\pi (s) - V_{P_{\mathrm{sim}}}^\pi (s) |\right] + \frac{\gamma R_{\mathrm{max}}}{(1 - \gamma)} \E_{q, \pi} \left[ \sum_{s' \in \StateRL} \Big| P_{\mathrm{real}} (s' |s,a) -P_{\mathrm{sim}} (s' | s, a) \Big| \right] \label{eq_3}
\end{align}
by applying the Bellman equation $V^\pi (s) = \sum_{a \in \mathcal{A}} \pi (a| s) \big( R (s, a) + \gamma \sum_{s' \in \StateRL} P (s' | s, a) V^\pi (s') \big)$ to both $V_{P_{\mathrm{real}}}^\pi (s)$ and $V_{P_{\mathrm{sim}}}^\pi (s)$ to obtain \eqref{eq_1}, by summing and subtracting $\gamma \E_{q,\pi} [ \sum_{s' \in \StateRL} V_{P_{\mathrm{sim}}}^\pi P_{\mathrm{real}} (s' | s,a)]$ to obtain \eqref{eq_2}, by exploiting $|a + b| \leq |a| + |b|$, $\sum_{s' \in \StateRL} P_{\mathrm{real}} (s'|s, a) = 1$, $V^\pi_{P_\mathrm{sim}} (s) \leq R_{\mathrm{max}} / (1 - \gamma)$, and stationarity of $q$ to obtain \eqref{eq_3}, from which the result follows.

{\color{black}

\subsection{Proof of Theorem~\ref{thm:gradient_analysis}}
\label{app:gradient_analysis_proof}

\begin{proof}
Throughout the proof, $\theta$ and the target network are fixed, and all gradients are taken with respect to $\theta$ with $V^T$ treated as a constant. We assume that $\Preal$ is absolutely continuous w.r.t. $\Psim$, so that the density ratio $w$ is well defined, and that all the second moments appearing below are finite.

\textbf{Notation.} A sample of $\tilde D_{\mathrm{real}}$ is $x = (s, a, s', \tilde s'_1, \dots, \tilde s'_K)$, where $(s,a) \sim \Preal$, $s' \sim P_{\mathrm{real}}(s,a)$, and $\tilde s'_1, \dots, \tilde s'_K$ are drawn i.i.d.\ from $P_{\mathrm{sim}}(s,a)$, independently of $s'$ given $(s,a)$. A sample of $D_{\mathrm{sim}}$ is $\tilde x = (\tilde s, \tilde a, \tilde s')$, where $(\tilde s, \tilde a) \sim \Psim$ and $\tilde s' \sim P_{\mathrm{sim}}(\tilde s, \tilde a)$. In~\eqref{eq:cv_grad}, the simulated Bellman error is computed with the empirical average of the simulated next-state values,
\begin{equation*}
    \delta^{\mathrm{sim}}_\theta(x) := Q_\theta(s,a) - \big(r(s,a) + \gamma\, \bar V_K(x)\big),
    \qquad
    \bar V_K(x) := \frac{1}{K}\sum_{k=1}^K V^T(\tilde s'_k),
\end{equation*}
and $\delta^{\mathrm{sim}}_\theta(\tilde x)$ is defined analogously with the single next state $\tilde s'$ (i.e., $K = 1$). We also denote the expected next-state values as
\begin{equation*}
    \mu_{\mathrm{real}}(s,a) := \E_{s' \sim P_{\mathrm{real}}(s,a)}\big[V^T(s')\big],
    \qquad
    \mu_{\mathrm{sim}}(s,a) := \E_{\tilde s' \sim P_{\mathrm{sim}}(s,a)}\big[V^T(\tilde s')\big],
\end{equation*}
so that $b(s,a) = 2\big(Q_\theta(s,a) - r(s,a) - \gamma\,\mu_{\mathrm{real}}(s,a)\big)\nabla_\theta Q_\theta(s,a)$, $\tilde b(s,a) = 2\big(Q_\theta(s,a) - r(s,a) - \gamma\,\mu_{\mathrm{sim}}(s,a)\big)\nabla_\theta Q_\theta(s,a)$, and $\Delta^2_{V^T} = \E_{s,a\sim\Preal}\big[\big(\mu_{\mathrm{real}}(s,a) - \mu_{\mathrm{sim}}(s,a)\big)^2\big]$.

\textbf{Preliminary: conditional expectations of the gradients.} Since $V^T$ is treated as a constant, $\nabla_\theta \delta^{\mathrm{real}}_\theta(x) = \nabla_\theta \delta^{\mathrm{sim}}_\theta(x) = \nabla_\theta Q_\theta(s,a)$, and therefore
\begin{equation}
    \nabla_\theta \delta^{\mathrm{real}}_\theta(x)^2 = 2\,\delta^{\mathrm{real}}_\theta(x)\,\nabla_\theta Q_\theta(s,a),
    \qquad
    \nabla_\theta \delta^{\mathrm{sim}}_\theta(x)^2 = 2\,\delta^{\mathrm{sim}}_\theta(x)\,\nabla_\theta Q_\theta(s,a).
    \label{eq:pf_grad_linear}
\end{equation}
Both gradients are linear in the corresponding TD error. Since $Q_\theta(s,a)$, $r(s,a)$ and $\nabla_\theta Q_\theta(s,a)$ are fixed given $(s,a)$, taking conditional expectations in~\eqref{eq:pf_grad_linear} gives
\begin{align}
    \E\big[\nabla_\theta \delta^{\mathrm{real}}_\theta(x)^2 \,\big|\, s,a\big]
    &= 2\big(Q_\theta(s,a) - r(s,a) - \gamma\,\E[V^T(s') \mid s,a]\big)\nabla_\theta Q_\theta(s,a)
    = b(s,a), \label{eq:pf_cond_real} \\
    \E\big[\nabla_\theta \delta^{\mathrm{sim}}_\theta(x)^2 \,\big|\, s,a\big]
    &= 2\big(Q_\theta(s,a) - r(s,a) - \gamma\,\E[\bar V_K(x) \mid s,a]\big)\nabla_\theta Q_\theta(s,a)
    = \tilde b(s,a), \label{eq:pf_cond_sim}
\end{align}
where~\eqref{eq:pf_cond_sim} uses $\E[\bar V_K(x) \mid s,a] = \mu_{\mathrm{sim}}(s,a)$, which holds for every $K \geq 1$. In particular, \eqref{eq:pf_cond_sim} also holds for the samples of $D_{\mathrm{sim}}$ ($K = 1$), i.e., $\E\big[\nabla_\theta \delta^{\mathrm{sim}}_\theta(\tilde x)^2 \,\big|\, \tilde s, \tilde a\big] = \tilde b(\tilde s, \tilde a)$.

\textbf{Proof of (i).} Since the samples within each dataset are i.i.d., by linearity of expectation we write
\begin{align}
    \E[\hat g_\lambda]
    &= \E_{x}\big[\nabla_\theta \delta^{\mathrm{real}}_\theta(x)^2 - \lambda\,\nabla_\theta \delta^{\mathrm{sim}}_\theta(x)^2\big]
    + \lambda\,\E_{\tilde x}\big[w(\tilde s, \tilde a)\,\nabla_\theta \delta^{\mathrm{sim}}_\theta(\tilde x)^2\big] \nonumber \\
    &= \E_{s,a\sim\Preal}\big[b(s,a) - \lambda\,\tilde b(s,a)\big]
    + \lambda\,\E_{s,a\sim\Psim}\big[w(s,a)\,\tilde b(s,a)\big] \label{eq:pf_i_1} \\
    &= \E_{s,a\sim\Preal}\big[b(s,a)\big] - \lambda\,\E_{s,a\sim\Preal}\big[\tilde b(s,a)\big]
    + \lambda\,\E_{s,a\sim\Preal}\big[\tilde b(s,a)\big] \label{eq:pf_i_2} \\
    &= \E_{s,a\sim\Preal}\big[b(s,a)\big], \nonumber
\end{align}
by applying the tower property together with~\eqref{eq:pf_cond_real}--\eqref{eq:pf_cond_sim} to obtain~\eqref{eq:pf_i_1}, and the definition of the density ratio to obtain~\eqref{eq:pf_i_2} as
\begin{equation*}
    \E_{s,a\sim\Psim}\big[w(s,a)\,\tilde b(s,a)\big] = \sum_{(s,a)} \Psim(s,a)\, \frac{\Preal(s,a)}{\Psim(s,a)}\, \tilde b(s,a) = \E_{s,a\sim\Preal}\big[\tilde b(s,a)\big].
\end{equation*}
The right-hand side does not depend on $\lambda$, hence $\E[\hat g_\lambda] = \E[\hat g_0]$. Finally, by~\eqref{eq:pf_grad_linear} and the tower property, $\E_{s,a\sim\Preal}[b(s,a)] = \E_{\pi, P_{\mathrm{real}}}\big[2\,\delta_\theta(s,a,s')\,\nabla_\theta Q_\theta(s,a)\big] = \nabla_\theta \E_{\pi, P_{\mathrm{real}}}\big[\delta_\theta(s,a,s')^2\big]$, which is the gradient of the TD objective~\eqref{TD objective} with the target network held fixed.

\textbf{Proof of (ii).} We write $\hat g_\lambda = \frac{1}{n}\sum_{i=1}^n u(x_i) + \frac{\lambda}{N}\sum_{j=1}^N v(\tilde x_j)$, where
\begin{equation*}
    u(x) := \nabla_\theta \delta^{\mathrm{real}}_\theta(x)^2 - \lambda\,\nabla_\theta \delta^{\mathrm{sim}}_\theta(x)^2,
    \qquad
    v(\tilde x) := w(\tilde s, \tilde a)\,\nabla_\theta \delta^{\mathrm{sim}}_\theta(\tilde x)^2.
\end{equation*}
The vectors $u(x_1), \dots, u(x_n), v(\tilde x_1), \dots, v(\tilde x_N)$ are independent, the $u(x_i)$ are identically distributed, and so are the $v(\tilde x_j)$. Hence, all the cross terms $\E\langle u(x_i) - \E u,\, u(x_{i'}) - \E u\rangle$ for $i \neq i'$, $\E\langle v(\tilde x_j) - \E v,\, v(\tilde x_{j'}) - \E v\rangle$ for $j \neq j'$, and $\E\langle u(x_i) - \E u,\, v(\tilde x_j) - \E v\rangle$ vanish, and
\begin{equation}
    \mathbb{V}[\hat g_\lambda]
    = \frac{1}{n}\,\mathbb{V}[u(x)] + \frac{\lambda^2}{N}\,\mathbb{V}[v(\tilde x)]
    = \frac{1}{n}\,\mathbb{V}[u(x)] + \frac{\lambda^2}{N}\,\sigma^2_{\IS},
    \label{eq:pf_ii_split}
\end{equation}
where the last equality follows from the definition of $\sigma^2_{\IS}$. It remains to compute $\mathbb{V}[u(x)]$. By the law of total variance,
\begin{align}
    \mathbb{V}[u(x)]
    &= \E_{s,a\sim\Preal}\Big[\big\|\E[u(x) \mid s,a] - \E[u(x)]\big\|_2^2\Big] \nonumber \\
    &\quad + \E_{s,a\sim\Preal}\Big[\E\big[\|u(x) - \E[u(x) \mid s,a]\|_2^2 \,\big|\, s,a\big]\Big],
    \label{eq:pf_ii_total}
\end{align}
where the cross term vanishes since $\E\big[u(x) - \E[u(x) \mid s,a] \,\big|\, s,a\big] = 0$.

For the first term of~\eqref{eq:pf_ii_total}, by~\eqref{eq:pf_cond_real}--\eqref{eq:pf_cond_sim} we have $\E[u(x) \mid s,a] = b(s,a) - \lambda\,\tilde b(s,a)$ and $\E[u(x)] = \E_{s,a\sim\Preal}[b(s,a) - \lambda\,\tilde b(s,a)]$, so that the first term equals $\mathbb{V}[b - \lambda\tilde b]$, with $b, \tilde b$ evaluated at $(s,a) \sim \Preal$.

For the second term of~\eqref{eq:pf_ii_total}, by~\eqref{eq:pf_grad_linear} and the definition of the TD errors, we write
\begin{align}
    u(x) - \E[u(x) \mid s,a]
    &= 2\big(\delta^{\mathrm{real}}_\theta(x) - \E[\delta^{\mathrm{real}}_\theta(x) \mid s,a]\big)\nabla_\theta Q_\theta(s,a) \nonumber \\
    &\quad - 2\lambda\big(\delta^{\mathrm{sim}}_\theta(x) - \E[\delta^{\mathrm{sim}}_\theta(x) \mid s,a]\big)\nabla_\theta Q_\theta(s,a) \nonumber \\
    &= -2\gamma\Big(\big(V^T(s') - \mu_{\mathrm{real}}(s,a)\big) - \lambda\big(\bar V_K(x) - \mu_{\mathrm{sim}}(s,a)\big)\Big)\nabla_\theta Q_\theta(s,a).
    \label{eq:pf_ii_noise}
\end{align}
Taking the squared norm of~\eqref{eq:pf_ii_noise} and the conditional expectation, we obtain
\begin{align}
    &\E\big[\|u(x) - \E[u(x) \mid s,a]\|_2^2 \,\big|\, s,a\big] \nonumber \\
    &\quad= 4\gamma^2\,\|\nabla_\theta Q_\theta(s,a)\|_2^2\;
    \E\Big[\big(\big(V^T(s') - \mu_{\mathrm{real}}(s,a)\big) - \lambda\big(\bar V_K(x) - \mu_{\mathrm{sim}}(s,a)\big)\big)^2 \,\Big|\, s,a\Big] \nonumber \\
    &\quad= 4\gamma^2\,\|\nabla_\theta Q_\theta(s,a)\|_2^2\,
    \Big(\mathbb{V}_{s' \sim P_{\mathrm{real}}(s,a)}\big[V^T(s')\big] + \lambda^2\,\mathbb{V}\big[\bar V_K(x) \mid s,a\big]\Big) \label{eq:pf_ii_indep} \\
    &\quad= 4\gamma^2\,\|\nabla_\theta Q_\theta(s,a)\|_2^2\,
    \Big(\mathbb{V}_{s' \sim P_{\mathrm{real}}(s,a)}\big[V^T(s')\big] + \frac{\lambda^2}{K}\,\mathbb{V}_{\tilde s' \sim P_{\mathrm{sim}}(s,a)}\big[V^T(\tilde s')\big]\Big), \label{eq:pf_ii_K}
\end{align}
where we obtain~\eqref{eq:pf_ii_indep} since $s'$ and $\tilde s'_1, \dots, \tilde s'_K$ are independent given $(s,a)$, so that the cross term vanishes,
\begin{align*}
    &\E\big[(V^T(s') - \mu_{\mathrm{real}}(s,a))(\bar V_K(x) - \mu_{\mathrm{sim}}(s,a)) \,\big|\, s,a\big] \\
    &\quad= \E\big[V^T(s') - \mu_{\mathrm{real}}(s,a) \,\big|\, s,a\big]\, \E\big[\bar V_K(x) - \mu_{\mathrm{sim}}(s,a) \,\big|\, s,a\big] = 0,
\end{align*}
and we obtain~\eqref{eq:pf_ii_K} since $\bar V_K(x)$ is the average of $K$ i.i.d.\ samples of $V^T(\tilde s')$ with $\tilde s' \sim P_{\mathrm{sim}}(s,a)$. Taking the expectation of~\eqref{eq:pf_ii_K} over $(s,a) \sim \Preal$, the second term of~\eqref{eq:pf_ii_total} equals $\sigma_T^2 + \lambda^2\tilde\sigma_T^2/K$ by the definitions of $\sigma_T^2$ and $\tilde\sigma_T^2$. Plugging both terms into~\eqref{eq:pf_ii_total}, and then into~\eqref{eq:pf_ii_split}, gives (ii).

\textbf{Proof of (iii).} By expanding the squared norm in the definition of the variance, we have
\begin{equation*}
    \mathbb{V}[b - \lambda\tilde b]
    = \E\big\|(b - \E b) - \lambda(\tilde b - \E\tilde b)\big\|_2^2
    = \mathbb{V}[b] - 2\lambda\,\mathrm{Cov}(b, \tilde b) + \lambda^2\,\mathbb{V}[\tilde b].
\end{equation*}
Plugging it into (ii) and denoting $C := \mathrm{Cov}(b, \tilde b)$ and $D := \mathbb{V}[\tilde b] + \tilde\sigma_T^2/K + n\,\sigma^2_{\IS}/N$, we obtain
\begin{equation}
    n\,\mathbb{V}[\hat g_\lambda] = \mathbb{V}[b] + \sigma_T^2 - 2\lambda\,C + \lambda^2 D.
    \label{eq:pf_iii_quad}
\end{equation}
If $D > 0$, the right-hand side of~\eqref{eq:pf_iii_quad} is a strictly convex quadratic function of $\lambda$. Setting its derivative $-2C + 2\lambda D$ to zero gives the minimizer $\lambda^\star = C/D$, and substituting it back into~\eqref{eq:pf_iii_quad} gives
\begin{equation}
    n\,\mathbb{V}[\hat g_{\lambda^\star}] = \mathbb{V}[b] + \sigma_T^2 - \frac{C^2}{D}.
    \label{eq:pf_iii_opt}
\end{equation}
Setting $\lambda = 0$ in~\eqref{eq:pf_iii_quad} gives $n\,\mathbb{V}[\hat g_0] = \mathbb{V}[b] + \sigma_T^2$, the variance of real-only TD. Since $C^2/D \geq 0$, we conclude $\mathbb{V}[\hat g_{\lambda^\star}] \leq \mathbb{V}[\hat g_0]$, with equality if and only if $C = 0$. If instead $D = 0$, then $\mathbb{V}[\tilde b] = 0$ and, by the Cauchy--Schwarz inequality, $|C| \leq \sqrt{\mathbb{V}[b]\,\mathbb{V}[\tilde b]} = 0$, so that~\eqref{eq:pf_iii_quad} does not depend on $\lambda$ and the claim holds trivially (with $\lambda^\star := 0$). Finally, the Cauchy--Schwarz inequality also gives $C^2 \leq \mathbb{V}[b]\,\mathbb{V}[\tilde b] \leq \mathbb{V}[b]\,D$, so that~\eqref{eq:pf_iii_opt} is bounded from below by $\sigma_T^2$: no choice of $\lambda$ reduces the variance below the real transition noise.

\textbf{Proof of (iv).} We write $b - \lambda\tilde b = (1 - \lambda)\,b + \lambda\,(b - \tilde b)$. For any random vectors $X, Y$, applying $\|x + y\|_2^2 \leq 2\|x\|_2^2 + 2\|y\|_2^2$ to $x = X - \E X$ and $y = Y - \E Y$ and taking expectations gives $\mathbb{V}[X + Y] \leq 2\,\mathbb{V}[X] + 2\,\mathbb{V}[Y]$. With $X = (1-\lambda)\,b$ and $Y = \lambda\,(b - \tilde b)$, we get
\begin{equation}
    \mathbb{V}[b - \lambda\tilde b] \leq 2(1-\lambda)^2\,\mathbb{V}[b] + 2\lambda^2\,\mathbb{V}[b - \tilde b].
    \label{eq:pf_iv_1}
\end{equation}
At each $(s,a)$, the terms $Q_\theta(s,a)$, $r(s,a)$ and $\nabla_\theta Q_\theta(s,a)$ are the same in $b$ and $\tilde b$, so that they cancel in the difference, giving
\begin{equation}
    b(s,a) - \tilde b(s,a) = 2\gamma\,\big(\mu_{\mathrm{sim}}(s,a) - \mu_{\mathrm{real}}(s,a)\big)\,\nabla_\theta Q_\theta(s,a).
    \label{eq:pf_iv_2}
\end{equation}
Then, we write
\begin{align}
    \mathbb{V}[b - \tilde b]
    &\leq \E_{s,a\sim\Preal}\big\|b(s,a) - \tilde b(s,a)\big\|_2^2 \label{eq:pf_iv_3} \\
    &= 4\gamma^2\,\E_{s,a\sim\Preal}\Big[\big(\mu_{\mathrm{real}}(s,a) - \mu_{\mathrm{sim}}(s,a)\big)^2\,\|\nabla_\theta Q_\theta(s,a)\|_2^2\Big] \label{eq:pf_iv_4} \\
    &\leq 4\gamma^2 G^2\,\E_{s,a\sim\Preal}\Big[\big(\mu_{\mathrm{real}}(s,a) - \mu_{\mathrm{sim}}(s,a)\big)^2\Big]
    = 4\gamma^2 G^2\,\Delta^2_{V^T}, \label{eq:pf_iv_5}
\end{align}
by exploiting $\mathbb{V}[X] = \E\|X\|_2^2 - \|\E X\|_2^2 \leq \E\|X\|_2^2$ to obtain~\eqref{eq:pf_iv_3}, by~\eqref{eq:pf_iv_2} and $\|c\,z\|_2^2 = c^2\|z\|_2^2$ for a scalar $c$ to obtain~\eqref{eq:pf_iv_4}, and by $\|\nabla_\theta Q_\theta(s,a)\|_2 \leq G$ for all $(s,a)$, together with the definition of $\Delta^2_{V^T}$, to obtain~\eqref{eq:pf_iv_5}. Plugging~\eqref{eq:pf_iv_5} into~\eqref{eq:pf_iv_1} gives (iv).
\end{proof}

}

\subsection{Proof of Unbiased Gradients with Masking}
\label{app:mask_bias_proof}
\begin{proof}
For the masked loss version we have
\[
\E[\hat g_\lambda]
= \E\!\left[
\nabla_\theta \delta_{\theta}^{\mathrm{real}}(x)^2
- \lambda \mathbb{I}_m \nabla_\theta \delta_{\theta}^{\mathrm{sim}}(x)^2
\right]
+ \E\!\left[
\lambda \mathbb{I}_m w(\tilde s,\tilde a)\nabla_\theta \delta_{\theta}^{\mathrm{sim}}(\tilde x)^2
\right].
\]
By definition of the density ratio $w(s,a)$, we have
\[
\E_{\tilde x \sim \Psim}\!\left[
w(\tilde s,\tilde a) \mathbb{I}_m \nabla_\theta \delta_{\theta}^{\mathrm{sim}}(\tilde x)^2
\right]
=
\E_{x \sim \Preal}\!\left[
\mathbb{I}_m \nabla_\theta \delta_{\theta}^{\mathrm{sim}}(x)^2
\right].
\]
Therefore, the simulator terms cancel, yielding
\[
\E[\hat g_\lambda]
=
\E\!\left[
\nabla_\theta \delta_{\theta}^{\mathrm{real}}(x)^2
\right].
\]
This is exactly the gradient of the squared TD error objective, implying that $\hat g_\lambda$ is an unbiased estimator.
\end{proof}

\section{Robustness to Simulation Bias Experiment}
\label{app:robustness}
\begin{figure*}[!h]
\centering
    \includegraphics[width=1.0\textwidth]{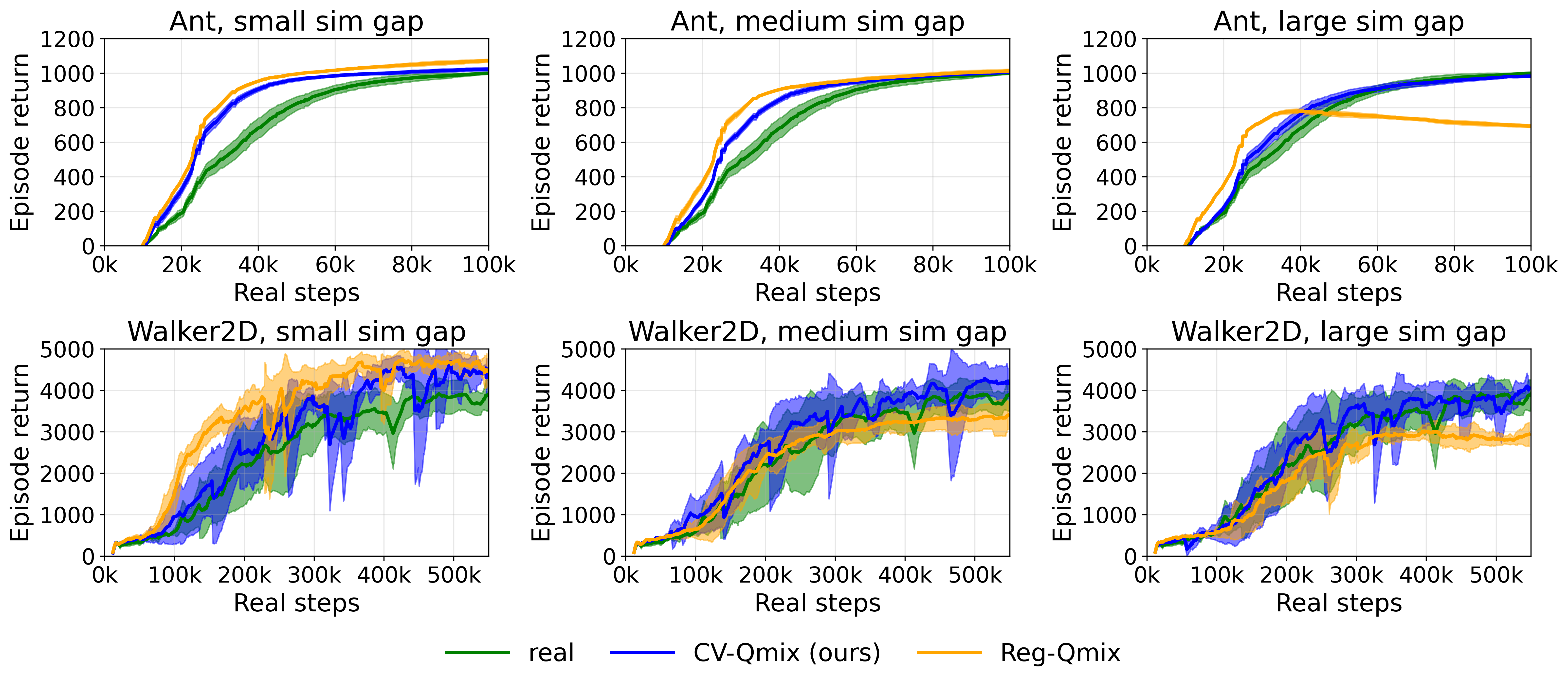}
    \caption{\textbf{Policy optimization robustness to simulation bias.} Comparison of data-mixing strategies under increasing dynamics mismatch for Ant and Walker2D tasks. Simulator parameters (friction, joint armature, joint damping) are perturbed relative to the real environment. Showing online optimization (mean $\pm$ std over 5 seeds).}
    \label{fig online different distance}
\end{figure*}

We study the behavior of simulation mixing methods across a range of \emph{sim-to-real gaps}, by progressively increasing the deviation of simulator parameters from the real environment. To this end, we use the online SAC variant in which, for both mixing methods (Reg-Qmix and CV-Qmix), we conservatively perform one policy update per real environment interaction, while incorporating a large number of additional interactions using simulated data.

In these experiments, we additionally evaluate the MuJoCo Ant and Walker2D tasks. For the Ant task, we reduce the motor strength in the real environment to $30\%$ of its default value. In our testing we observed that lowering motor strength of Ant amplifies the effect of dynamics mismatches, making performance degradation under sim-to-real transfer more pronounced, creating a meaningful bias for evaluation.

In this experiment, we vary joint damping, armature, and joint friction to construct simulators with progressively increasing dynamics mismatch relative to \( M_{\mathrm{real}} \) (from left to right in Fig.~\ref{fig online different distance}). For exact experiments setup details see appendix \ref{app: sim parameters}.

Across both tasks, when \( M_{\mathrm{sim}} \) is close to \( M_{\mathrm{real}} \), both regular mixing and CV mixing improve policy performance relative to training on real data alone. However, as the dynamics mismatch increases, regular mixing degrades performance, whereas CV mixing remains comparable to the real-only baseline. These results highlight the robustness of CV mixing, which mitigates the introduction of simulator-induced bias that can otherwise destabilize the optimization process.

\newpage
\section{Experiments Setup}
\subsection{Training Hyper parameters}
\label{app:Training parameters}

\begin{table*}[h]
\centering
\scriptsize
\setlength{\tabcolsep}{3pt}
\renewcommand{\arraystretch}{0.9}

\begin{minipage}[t]{0.45\textwidth}
\centering

\captionof{table}{Walker2D and Ant robustness experiment Hyperparameters}
\begin{tabular}{lcc}
\hline
Hyperparameter & Walker2D-v5 & Ant-v5 \\
\hline
critic learning rate & 0.0003 & 0.0003 \\
policy learning rate & 0.0003 & 0.0003 \\
real step frequency & 1 & 1 \\
batch size real & 128 & 128 \\
batch size sim & 128 & 128 \\
replay buffer size real & 100000 & 100000 \\
replay buffer size sim & 11000000 & 11000000 \\
sim step frequency & 1 & 1 \\
num steps in sim & 10 & 10 \\
critic update frequency & 1 & 1 \\
target network update frequency & 1 & 1 \\
$\tau$ & 0.005 & 0.005 \\
$\gamma$ & 0.99 & 0.99 \\
policy update frequency & 2 & 2 \\
policy update steps & 2 & 2 \\
classifier update frequency & 10 & 10 \\
classifier update steps & 10 & 10 \\
classifier learning rate & 0.0003 & 0.0003 \\
classifier batch size & 128 & 128 \\
initial $\lambda$ & 0 & 0 \\
$\lambda$ decay & 0.001 & 0.001 \\
SAC $\alpha$ & 0.05 & 0.2 \\
CV mask $\beta$ & 3.0 & 3.0 \\
\hline
\end{tabular}

\vspace{1em}
\captionof{table}{Hopper and Sliding sample efficiency online SAC experiments Hyperparameters}
\begin{tabular}{lc}
\hline
Hyperparameter & Value \\
\hline
critic learning rate & 0.0003 \\
policy learning rate & 0.0003 \\
real step frequency & 10 \\
batch size real & 128 \\
batch size sim & 128 \\
replay buffer size real & 100000 \\
replay buffer size sim & 11000000 \\
sim step frequency & 1 \\
num steps in sim & 1 \\
critic update frequency & 1 \\
target network update frequency & 1 \\
$\tau$ & 0.005 \\
$\gamma$ & 0.99 \\
policy update frequency & 2\\
policy update steps & 2 \\
classifier update frequency & 10\\
classifier update steps & 10\\
classifier learning rate & 0.0003\\
classifier batch size & 128 \\
initial $\lambda$ & 0 \\
$\lambda$ decay & 0.001 \\
SAC $\alpha$ & 0.05 \\
CV mask $\beta$ & 3.0 \\
\hline
\end{tabular}
\end{minipage}
\hfill
\begin{minipage}[t]{0.45\textwidth}
\centering

\captionof{table}{Sliding: Behavioral Policy Training in Simulation}
\begin{tabular}{lc}
\hline
Hyperparameter & Value \\
\hline
critic learning rate & 0.0003 \\
policy learning rate & 0.0003 \\
SAC $\alpha$ & 0.05 \\
batch size & 128 \\
replay buffer size & 100000 \\
critic update frequency & 1 \\
target network update frequency & 1 \\
$\tau$ & 0.005 \\
$\gamma$ & 0.99 \\
policy update frequency & 2 \\
policy update steps & 2 \\
total steps & 300000 \\
\hline
\end{tabular}

\captionof{table}{Hopper: Behavioral Policy Training in Simulation}
\begin{tabular}{lc}
\hline
Hyperparameter & Value \\
\hline
critic learning rate & 0.0003 \\
policy learning rate & 0.0003 \\
SAC $\alpha$ & 0.2 \\
batch size & 128 \\
replay buffer size & 100000 \\
critic update frequency & 1 \\
target network update frequency & 1 \\
$\tau$ & 0.005 \\
$\gamma$ & 0.99 \\
policy update frequency & 2 \\
policy update steps & 2 \\
total steps & 1000000 \\
\hline
\end{tabular}

\captionof{table}{Offline CQL Training Hyperparameters}
\begin{tabular}{lcc}
\hline
Hyperparameter & Sliding & Hopper \\
\hline
critic learning rate & 0.0001 & 0.0001 \\
policy learning rate & 0.0003 & 0.0003 \\
batch size real & 128 & 128 \\
batch size sim & 128 & 128 \\
replay buffer size real & 20000 & 2000 \\
replay buffer size sim & 1000000 & 100000 \\
critic update frequency & 1 & 1 \\
target network update frequency & 1 & 1 \\
$\tau$ & 0.005 & 0.005 \\
$\gamma$ & 0.99 & 0.99 \\
policy update frequency & 2 & 2 \\
policy update steps & 2 & 2 \\
classifier subset split & 0.3 & 0.3 \\
classifier training steps & 1000 & 1000 \\
classifier learning rate & 0.0003 & 0.0003 \\
initial $\lambda$ & 0 & 0\\
$\lambda$ decay & 0.001 & 0.001\\
SAC $\alpha$ & auto-tune & 0.2 \\
CQL $\alpha$ & 0.5 & 0.5 \\
maximum CQL penalty gap & 20 & 20 \\
CV mask $\beta$ & 3.0 & 3.0 \\
\hline
\end{tabular}

\vspace{1em}


\end{minipage}
\end{table*}

\newpage
\subsection{Off-Dynamics finetuning benchmark Hyper parameters}

\begin{table*}[h]
\centering
\scriptsize
\setlength{\tabcolsep}{3pt}
\renewcommand{\arraystretch}{0.9}

\begin{minipage}[t]{0.45\textwidth}
\centering
\captionof{table}{Real-only (SAC fine-tuning) Hyperparameters}
\begin{tabular}{lcc}
\toprule
Hyperparameter & Walker2D-v5 & Ant-v5 \\
\midrule
critic learning rate             & 0.0003  & 0.0003  \\
policy learning rate             & 0.0003  & 0.0003  \\
total real env steps             & 100000 & 200000  \\
real step frequency              & 1       & 1       \\
batch size real                  & 128     & 128     \\
replay buffer size real          & 100000 & 100000 \\
learning starts                  & 25000   & 5000    \\
critic update frequency          & 1       & 1       \\
target network update frequency  & 1       & 1       \\
$\tau$                           & 0.005   & 0.005   \\
$\gamma$                         & 0.99    & 0.99    \\
policy update frequency          & 2       & 2       \\
policy update steps              & 2       & 2       \\
SAC $\alpha$                     & auto-tuned & 0.2  \\
\bottomrule
\end{tabular}

\vspace{1em}

\captionof{table}{CV-Qmix Hyperparameters}
\begin{tabular}{lcc}
\toprule
Hyperparameter & Walker2D-v5 & Ant-v5 \\
\midrule
critic learning rate             & 0.0003  & 0.0003  \\
policy learning rate             & 0.0003  & 0.0003  \\
total update steps               & 1000000 & 1000000 \\
real step frequency              & 5       & 5       \\
sim step frequency               & 1       & 1       \\
num steps in sim                 & 100     & 100     \\
critic batch size real           & 64      & 64      \\
critic batch size sim            & 256     & 256     \\
policy batch size real           & 128     & 128     \\
policy batch size sim            & 64      & 64      \\
replay buffer size real          & 100000  & 100000  \\
replay buffer size sim           & 1000000 & 1000000 \\
learning starts                  & 25000   & 5000    \\
critic update frequency          & 1       & 1       \\
target network update frequency  & 1       & 1       \\
$\tau$                           & 0.005   & 0.005   \\
$\gamma$                         & 0.99    & 0.99    \\
policy update frequency          & 2       & 2       \\
policy update steps              & 2       & 2       \\
critic grad-norm clip            & 50000   & 50000   \\
classifier update frequency      & 1       & 1       \\
classifier update steps          & 1       & 1       \\
classifier learning rate         & 0.0003  & 0.0003  \\
classifier weight decay          & 0.2     & 0.2     \\
classifier batch size            & 128     & 128     \\
classifier hidden layers / dim   & 2 / 256 & 2 / 256 \\
initial $\lambda$                & 0       & 0       \\
$\lambda$ decay                  & 0.001   & 0.001   \\
sim importance-ratio clip        & [0, 3]  & [0, 3]  \\
next-action samples              & 10      & 10      \\
CV mask $\beta$                  & 3.0     & 3.0     \\
SAC $\alpha$                     & auto-tuned & 0.2  \\
\bottomrule
\end{tabular}

\end{minipage}
\hfill
\begin{minipage}[t]{0.45\textwidth}
\centering

\captionof{table}{Reg-Qmix Hyperparameters}
\begin{tabular}{lcc}
\toprule
Hyperparameter & Walker2D-v5 & Ant-v5 \\
\midrule
critic learning rate             & 0.0003  & 0.0003  \\
policy learning rate             & 0.0003  & 0.0003  \\
total update steps               & 1000000 & 1000000 \\
real step frequency              & 5       & 5       \\
sim step frequency               & 1       & 1       \\
num steps in sim                 & 100     & 100     \\
critic batch size real           & 128     & 128     \\
critic batch size sim            & 128     & 128     \\
policy batch size real           & 128     & 128     \\
policy batch size sim            & 64      & 64      \\
replay buffer size real          & 100000  & 100000  \\
replay buffer size sim           & 1000000 & 1000000 \\
learning starts                  & 5000    & 5000    \\
critic update frequency          & 1       & 1       \\
target network update frequency  & 1       & 1       \\
$\tau$                           & 0.005   & 0.005   \\
$\gamma$                         & 0.99    & 0.99    \\
policy update frequency          & 2       & 2       \\
policy update steps              & 2       & 2       \\
mixing coefficient $\alpha_{mix}$ & 0.5    & 0.5     \\
SAC $\alpha$                     & auto-tuned & 0.2  \\
\bottomrule
\end{tabular}

\vspace{1em}

\captionof{table}{PAR Hyperparameters}
\begin{tabular}{lcc}
\toprule
Hyperparameter & Walker2D-v5 & Ant-v5 \\
\midrule
critic learning rate             & 0.0003  & 0.0003  \\
policy learning rate             & 0.0003  & 0.0003  \\
total update steps               & 1000000 & 1000000 \\
real step frequency              & 5       & 5       \\
sim step frequency               & 1       & 1       \\
num steps in sim                 & 100     & 100     \\
critic batch size real           & 128     & 128     \\
critic batch size sim            & 128     & 128     \\
policy batch size real           & 64      & 64      \\
policy batch size sim            & 64      & 64      \\
replay buffer size real          & 100000  & 100000  \\
replay buffer size sim           & 1000000 & 1000000 \\
learning starts                  & 5000    & 5000    \\
critic update frequency          & 1       & 1       \\
target network update frequency  & 1       & 1       \\
$\tau$                           & 0.005   & 0.005   \\
$\gamma$                         & 0.99    & 0.99    \\
policy update frequency          & 2       & 2       \\
policy update steps              & 2       & 2       \\
encoder learning rate            & 0.0003  & 0.0003  \\
encoder hidden dim               & 256     & 256     \\
representation dim ($z_s$)       & 256     & 256     \\
encoder update frequency         & 200     & 200     \\
encoder update steps             & 1       & 1       \\
encoder batch size               & 128     & 128     \\
reward penalty coefficient $\beta$ & 1.0   & 1.0     \\
SAC $\alpha$                     & auto-tuned & 0.2  \\
\bottomrule
\end{tabular}

\end{minipage}

\end{table*}

\begin{table}[h]
\scriptsize
\centering
\caption{VGDF Hyperparameters}
\begin{tabular}{lcc}
\toprule
Hyperparameter & Walker2D-v5 & Ant-v5 \\
\midrule
critic learning rate             & 0.0003  & 0.0003  \\
policy learning rate             & 0.0003  & 0.0003  \\
total update steps               & 1000000 & 1000000 \\
real step frequency              & 5       & 5       \\
sim step frequency               & 1       & 1       \\
num steps in sim                 & 100     & 100     \\
batch size real                  & 128     & 128     \\
batch size sim                   & 128     & 128     \\
policy batch size                & 128     & 128     \\
replay buffer size real          & 100000  & 100000  \\
replay buffer size sim           & 1000000 & 1000000 \\
learning starts                  & 5000    & 5000    \\
critic update frequency          & 1       & 1       \\
target network update frequency  & 1       & 1       \\
$\tau$                           & 0.005   & 0.005   \\
$\gamma$                         & 0.99    & 0.99    \\
policy update frequency          & 2       & 2       \\
policy update steps              & 2       & 2       \\
actor-critic grad-norm clip      & 100     & 100     \\
dynamics ensemble size / elites  & 7 / 5   & 7 / 5   \\
dynamics hidden dim              & 200     & 200     \\
dynamics learning rate           & 0.001   & 0.001   \\
dynamics batch size              & 256     & 256     \\
dynamics train frequency         & 250     & 250     \\
dynamics holdout ratio           & 0.1     & 0.1     \\
likelihood gate threshold        & 0.75    & 0.75    \\
sim samples before gating        & 100000  & 100000  \\
optimistic exploration           & True    & True    \\
SAC $\alpha$                     & auto-tuned & 0.2  \\
\bottomrule
\end{tabular}
\end{table}

\newpage
\subsection{MuJoCo environment setting}
\label{app: sim parameters}
\begin{table*}[h]
\centering
\scriptsize
\setlength{\tabcolsep}{3pt}
\renewcommand{\arraystretch}{0.9}

\begin{minipage}[t]{0.45\textwidth}
\centering

\captionof{table}{Hopper Task Parameters}
\begin{tabular}{lcccc}
\hline
Parameter & Real & Small Gap & Medium Gap & Large Gap \\
\hline
gravity & -9.81 & -8.82 & -7.84 & -6.86 \\
floor friction & 1.0 & 0.8 & 0.8 & 0.7 \\
torso length & 0.2 & 0.24 & 0.24 & 0.26 \\
\hline
\end{tabular}

\vspace{1em}
\captionof{table}{Sliding Task Parameters}
\begin{tabular}{lcc}
\hline
Parameter & Real & Sim \\
\hline
puck damping & 0.2 & 0.24 \\
\hline
\end{tabular}

\vspace{1em}
\captionof{table}{Walker2D Finetuning Task Parameters}
\begin{tabular}{lcccc}
\toprule
Parameter & Sim & Real        & Real         & Real        \\
          &     & (Small Gap) & (Medium Gap) & (Large Gap) \\
\midrule
gravity tilt (deg) & 0   & 4    & 6   & 8    \\
foot friction    & 1.9 & 2.85 & 3.8 & 4.75 \\
\bottomrule
\end{tabular}

\vspace{1em}
\captionof{table}{Humanoid Policy Eval Task}
\begin{tabular}{lcc}
\toprule
Parameter & Sim & Real \\
\midrule
gravity vector & (0,0,-9.81)   & (-3, 0, -8)  \\
\bottomrule
\end{tabular}

\end{minipage}
\hfill
\begin{minipage}[t]{0.45\textwidth}
\centering

\captionof{table}{Walker2D Robustness Experiment Parameters}
\begin{tabular}{lcccc}
\hline
Parameter & Real & Small Gap & Medium Gap & Large Gap \\
\hline
joint damping & 1 & 1 & 1.5 & 2 \\
joint friction & 0.7 & 0.6 & 0.5 & 0.4 \\
\hline
\end{tabular}
\vspace{1em}
\captionof{table}{Ant Robustness Experiment Parameters}
\begin{tabular}{lcccc}
\hline
Parameter & Real & Small Gap & Medium Gap & Large Gap \\
\hline
joint armature & 3 & 2.5 & 2 & 1.5 \\
joint damping & 2 & 2 & 1.5 & 1.5 \\
joint friction & 0.4 & 0.5 & 0.8 & 1 \\
\hline
\end{tabular}

\vspace{1em}
\captionof{table}{Ant Finetuning Task Parameters}
\begin{tabular}{lcccc}
\toprule
Parameter & Real & Small Gap & Medium Gap & Large Gap \\
\midrule
gravity tilt (deg)    & 0   & -2.5  & -5   & -30 \\
floor friction      & 1.0 & 1.375 & 1.75 & 2.5 \\
ankle motor gear    & 150 & 145   & 140  & 130 \\
torso CoM x-offset  & 0   & 0.1   & 0.2  & 0.4 \\
\bottomrule
\end{tabular}

\vspace{1em}

\end{minipage}

\end{table*}

\section{Additional Implementation Details}
\label{app:lambda}
To improve the stability of the practical CV loss \eqref{CV empiric loss} We clipped the loss value so it cannot be negative. In our experiments this helps the performance of CV-Qmix.
$$
L_\mathrm{CV}^\mathrm{clip}(\theta) = \mathrm{clip}(L_\mathrm{CV}(\theta), 0, \infty)
$$

\section{The role of $\lambda$ in CV-Qmix}

\begin{table}[h]
\centering
\begin{tabular}{lcc}
\toprule
Sim Gap & Ant & Walker2D \\
\midrule
Small  & $0.37 \pm 0.02$ & $0.45 \pm 0.06$ \\
Medium & $0.21 \pm 0.01$ & $0.32 \pm 0.06$ \\
Large  & $0.08 \pm 0.02$ & $0.25 \pm 0.05$ \\
\bottomrule
\end{tabular}
\caption{$\hat{\lambda}$ values of CV-Qmix by simulation gap in the fine-tuning experiment.
For each seed, $\hat{\lambda}$ is averaged over the full training run; values are mean $\pm$ std over 5 seeds.}
\label{table:lambda-values}
\end{table}

The learned adaptive coefficient $\lambda$ also exhibited a clear relationship with the simulator mismatch (see table~\ref{table:lambda-values}). when the simulator is strongly correlated with the real environment, the method assigns it a larger weight, whereas under a large mismatch the simulator contribution is automatically suppressed. This supports the interpretation of $\lambda$ as both a diagnostic of real--sim correlation and a soft gating mechanism for simulated information.

\section{Compute Resources}
\label{Compute section}
For all experiments, we used two NVIDIA L40S GPUs. The computational requirements of our method are modest, and the experiments can be run on standard GPUs with as little as 4GB of VRAM. Each experiment reported in the paper required approximately one day to complete. 

\section{Puck Sliding Environment}
\label{sec:Puck-sliding}
\begin{figure*}[h]
\centering
    \includegraphics[width=0.8\textwidth]{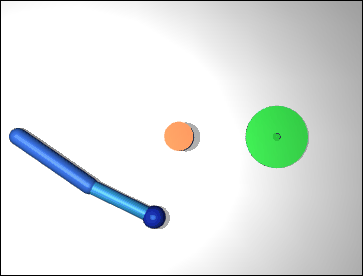}
    \caption{The environment consists of a two-link robotic arm that moves in the horizontal $(x,y)$ plane, along with a puck that can slide on the surface. The objective is for the arm to strike the puck so that it reaches a target location on the plane, which is placed far from the robot. The puck, arms joints, target positions are randomized at each episode.}
\end{figure*}

\end{document}